\documentclass[a4paper,11pt,oneside]{scrbook}  
\usepackage{amsmath}
\usepackage{adjustbox}
\usepackage{chngcntr}
\usepackage{caption}
\usepackage{paralist}
\usepackage{enumerate}
\usepackage{mathtools}
\usepackage{makecell}
\usepackage{hyperref}
\usepackage{multicol}
\usepackage{multirow}
\usepackage{graphicx}
\usepackage{layout} %
\usepackage{blindtext}  %
\usepackage{pgfgantt}
\usepackage{outlines}
\usepackage{float}
\usepackage{fancyhdr}
\usepackage[labelfont=bf]{caption}
\usepackage[utf8]{inputenc}
\usepackage{tabularx}
\usepackage{todonotes}
\usepackage[nolist]{acronym}
\usepackage[]{graphicx}    %
\newcommand{\ignore}[1]{}  %
\usepackage{rotating}
\usepackage[english]{babel}
\usepackage[utf8]{inputenc}
\usepackage{subfig}
\usepackage{booktabs}
\usepackage{url}
\usepackage{sidecap}
\usepackage{natbib}
\usepackage{hyperref,tablefootnote,footnotehyper}

\fancypagestyle{firstpage}{%
  \fancyhf{}%
  
  \cfoot{\tiny \textcopyright 2019 California Institute of Technology. Partial Government sponsorship acknowledged.}
}
\usepackage{xpatch}
\xapptocmd{\titlepage}{\thispagestyle{firstpage}}{}{}

\begin{document}

\begin{titlepage}
    \vspace*{-40mm}
    \enlargethispage{0.5cm}
	\centering
	
	\vspace{2.0cm}
	\includegraphics[width=0.5\textwidth]{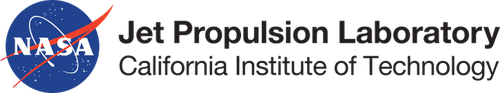}\par
	\vspace{0.5cm}
	{\Large \bfseries Final Report \par}
	{\Large \bfseries NASA NIAC Phase I Study \par}
	
	\vspace{0.5cm}
	
	{\huge  \scshape The Shapeshifter: a Morphing, Multi-Agent, Multi-Modal Robotic Platform for the Exploration of Titan (pre-print version) \par}
	
	\vspace{1.0cm}
	
	{\large Prepared for \par 
	Program Executive, NASA Innovative Advanced Concepts Program \par}

	\vspace{0.5cm}
	
	{\large Submitted May 31st, 2019}
	
	\vfill
	
	{\large Submitted by: \par
	Dr. Ali-akbar Agha-mohammadi, PI \par 
	\textit{Jet Propulsion Laboratory} \par 
	\textit{California Institute of Technology} \par
	{4800 Oak Grove Dr, Pasadena, CA 91109} \par
	{Phone: (626) 840-9140} \par 
	{Email: \textbf{aliagha@jpl.nasa.gov}} \par}
	
	\vspace{0.5cm}
	
	\newcolumntype{M}[1]{>{\centering\arraybackslash}m{#1}}
	\newcommand{\auth}[3][{\textit{Jet Propulsion Laboratory} \par \textit{California Institute of Technology}}]{#2 \par #1 \par \textbf{#3} \par}
	
	\noindent \begin{tabularx}{\textwidth}{M{5.5cm} M{5.5cm} M{5.5cm}}
	    \auth{Andrea Tagliabue }{atagliab@jpl.nasa.gov}
    & 
        \auth[{\textit{Dep. of Aeronautics and Astronautics} \par
        \textit{Stanford University}}]{Stephanie Schneider}{schneids@stanford.edu}
    &
        \auth{Dr. Benjamin Morrell}{benjamin.morrell@jpl.nasa.gov}
    \\
        \auth[{\textit{Dep. of Aeronautics and Astronautics} \par
        \textit{Stanford University}}]{Prof. Marco Pavone, Co-I}{pavone@stanford.edu}
    &
	    \auth{Dr. Jason Hofgartner, Co-I}{jason.d.hofgartner@jpl.nasa.gov}
	&
	    \auth{Dr. Issa A.D. Nesnas, Co-I}{nesnas@jpl.nasa.gov} 
	\\ 
	    \auth{Kalind Carpenter, Co-I}{kalind.c.carpenter@jpl.nasa.gov}
    &
        \auth{Dr. Rashied B. Amini, Co-I}{ramini@jpl.nasa.gov}
    & 
	    \auth{Dr. Arash Kalantari}{arash.kalantari@jpl.nasa.gov} 
	\\
	
		\auth{Dr. Alessandra Babuscia}{alessandra.babuscia@jpl.nasa.gov}
    &
        \auth[\textit{School of Engineering and Applied Sciences} \par \textit{University at Buffalo }]{Dr. Javid Bayandor}{bayandor@buffalo.edu}
    & 
	   \auth[\textit{Dep. of Astronomy} \par \textit{Cornell University}]{Prof. Jonathan Lunine}{jil45@cornell.edu} 
	\\

	\end{tabularx}

\end{titlepage}

\chapter*{Executive Summary}
\begin{figure}[h]
\centering
\includegraphics[width=\columnwidth]{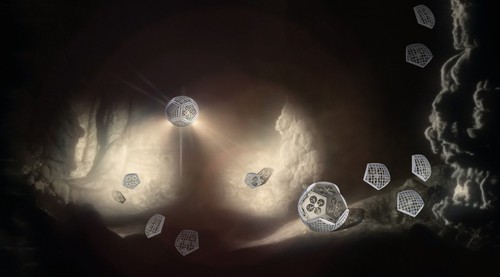}
      \caption{Artist's representation of the mission concept for the Shapeshifter. A Shapeshifter is constituted by smaller, flying robots that can combine together, morphing into a robot able to roll on the surface or swim in liquid basins. Image credits: Jose Mendez.}
\label{fig:intro_0}
\end{figure}

\addcontentsline{toc}{chapter}{\protect\numberline{}Executive Summary}
\counterwithout{figure}{section} %
In this report, we present a robotic platform, the Shapeshifter, that allows multi-domain and  redundant mobility on Saturn's moon Titan - and potentially other bodies with atmospheres. The Shapeshifter is a collection of simple and affordable robotic units, called Cobots, comparable to personal palm-size quadcopters. By attaching and detaching with each other, multiple Cobots can shape-shift into novel structures, capable of
\begin{inparaenum}[(a)]
\item rolling on a flat surface, to increase the traverse range,
\item flying in a flight array formation, and
\item swimming on or under liquid.
\end{inparaenum}
A ground station, called the Home-base, complements the robotic platform, hosting science instrumentation and providing power to recharge the batteries of the Cobots. An artist's concept of the platform, depicted in an exploration mission on Titan, is represented in Figure \ref{fig:intro_0}.
\par
Our Phase I study had the objective of providing an initial  assessment of  the  feasibility of the proposed robotic platform architecture, and in particular 
\begin{inparaenum}[(a)]
\item to characterize the expected science return of a mission to the Sotra-Patera region on Titan;
\item to verify the mechanical and algorithmic feasibility of building a multi-agent platform capable of flying, docking, rolling and un-docking;
\item to evaluate the increased range and efficiency of rolling on Titan w.r.t to flying;
\item to define a case-study of a mission for the exploration of the cryovolcano Sotra-Patera on Titan, whose expected variety of geological features challenges conventional mobility platforms.
\end{inparaenum}
The main results of our study can be summarized as follows:

\begin{itemize}

    \item \textbf{Science rationale:} Titan presents a compelling exploration target, being a laboratory to study organic synthesis processes, investigate complex organics that could be the building blocks of life, and a potential host for current life. A rich diversity of geological phenomena including lakes, caves, cryovolcanoes, dunes and planes also make Titan an ideal location to learn about the formation and evolution of planets and moons. These investigations require in-situ measurements in a range of extreme terrains, and Shapeshifter is specifically designed to enable sample gathering in these terrains. 
    \begin{figure}
        \centering
        \includegraphics[width=\linewidth]{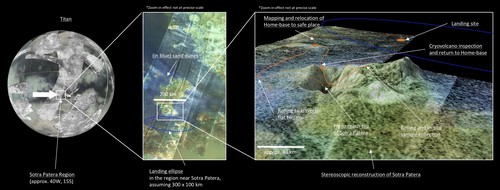}
          \caption{Example of mission scenario near Sotra Patera. The landing area is chosen in the proximity of the cryovolcano Sotra Patera, and we assume a landing ellipse of 100 km by 300 km. After landing, the Shapeshifter maps the surrounding environment and relocate the Home-base to a safe place. From there, the Shapeshifter moves, by rolling or flying, to the summit of the volcano, where in-situ samples are collected. After morphing into a Rollocopter to inspect the caves near the cavity of the volcano, Shapeshifter returns to the Home-base to analyze the collected samples and plan the next mission.}
        \label{fig:SotraPateraMissionArchitecture_0}
    \end{figure}

    \item \textbf{Shapeshifter platform principles:} We show that the Shapeshifter is constituted by multiple palm-sized multi-copters called Cobots, capable of docking and un-docking. By docking with each other, Cobots can morph into at least three different mobility configurations, which guarantee access to different environments. Such configurations are 
    \begin{inparaenum}[(a)]
        \item a sphere, capable of rolling on smooth surfaces for energy-efficient locomotion or to guarantee resilience to impacts in environments where high localization accuracy is not possible;
        \item a swarm of flying quadcopters or a flying array, capable of collecting spatiotemporal measurements or creating a mesh-network for communication in underground environments, such as caves and cryo-lava tubes;
        \item a torpedo, capable of swimming under Titan's mare to collect samples.
    \end{inparaenum}  
    We show that the Shapeshifter platform is also constituted by an Home-base, equivalent to a lander, which hosts scientific instruments, power (also used to recharge the Cobots) and communication equipment to relay data to Earth. 
    
    \item \textbf{Two-agents Shapeshifter prototype:} We experimentally show the flying, docking and rolling capabilities of a Shapeshifter constituted by two Cobots, presenting ad-hoc control algorithms. This is done by building a prototype based on two multi-copters enclosed in a hemicylindrical shell. Such shell allows the robots to dock together, creating a cylindrical structure which can roll on the ground by using the thrust force produced by its propellers. The two multi-copters can dock and un-dock via permanent-electromagnets placed in their frame. Multiple frames of the docking, rolling, un-docking maneuver performed with our prototype are shown in Figure \ref{fig:shapeshifter_movie_0}; the full video can be found in \cite{video}. We additionally derive and implement an attitude controller adequate to control the rolling speed of the prototype.
    \begin{figure}
        \centering
        \includegraphics[width=\linewidth]{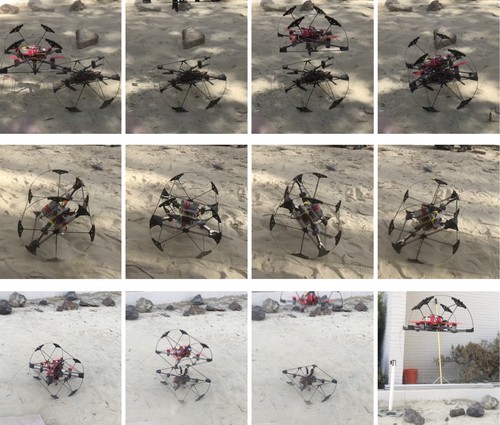}
        \caption{Frames from the video-clip \cite{video} of the experiments conducted with our two-Cobots Shapeshifter. \textit{Top:} Docking sequence. \textit{Center:} Rolling sequence. \textit{Bottom:} Un-docking sequence.}
        \label{fig:shapeshifter_movie_0}
    \end{figure}
   
    \item \textbf{Shapeshifter's energy-efficiency analysis:}  We  evaluate the energy-efficiency of the rolling-based mobility strategy by deriving an analytic model of the power consumption and by integrating it in a high-fidelity simulation environment, shown in Figure \ref{fig:gazebo_titan_0}. The simulator is based on ROS (Robotic Operative System) \cite{quigley2009ros} and Gazebo \cite{koenig2004design} and takes into account dynamics, aerodynamics, terramechanics and control algorithms. We compare the energy requirements of rolling and flying on Titan, showing that rolling can be up to two times more energy efficient than flying. Our analysis, as reported in Figure \ref{fig:multi_dim_0}, shows that rolling is more efficient than flying as long as the terrain is sufficiently compact (e.g. consolidated soil or hard sand) or sufficiently flat ($\approx0\deg$ or downhill). Our analysis additionally shows that, thanks to the high density and low gravity of Titan's atmosphere, even a small Cobot ($\approx 1$kg of mass, $\approx0.03$m$^3$ of volume and $\approx 2000$mAh of three-cell battery) can achieve high range ($\approx 100$km when flying and $\approx 200$km when rolling). The optimal range velocities, for a defined nominal case, are approximately $0.2$m/s for rolling and $1.7$m/s for flying, as shown in Figure \ref{fig:range_v_velocity_0}. We additionally investigate the increased benefit of flying w.r.t. to rolling as a function of the number of agents that compose a Shapeshifter, according to the best-case and worst-case analysis of the total aerodynamic drag of the platform as a function of the number of Cobots used in a rolling Shapeshifter (the Rollocopter) . We show that a Shapeshifter constituted by $3$ to $6$ agents may guarantee the best rolling range w.r.t the flying range.
    \begin{figure*} 
    \centering
    \subfloat[\small Advantage of rolling vs. flying, as it relates to surface slope and rolling resistance. The flying range is on the order of 130km for all terrain. The plot shows the difference between flying range and rolling range; add flying range to results shown here to get total rolling range.]{%
        \includegraphics[width=0.52\linewidth]{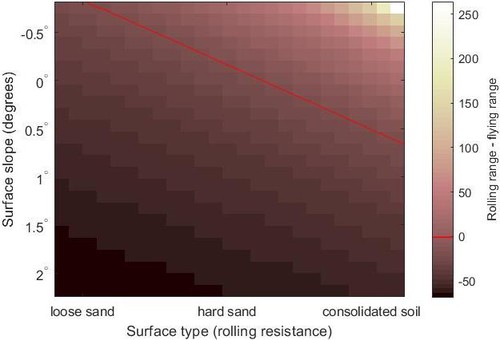}%
        \label{fig:multi_dim_0}%
        }%
    \hfill%
    \subfloat[\small Range vs. velocity for two Cobots on Titan, for flying and rolling along a surface of consolidated soil. Dashed lines indicate optimal velocity for each configuration.]{%
        \includegraphics[width=0.46\linewidth]{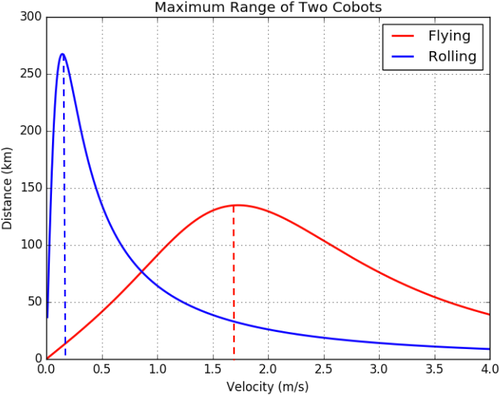}%
        \label{fig:range_v_velocity_0}%
        }%
    \caption{Results of the energy efficiency analysis for the two-agents Shapeshifter on Titan.}
    \end{figure*}
    
    \begin{figure}
        \centering
        \includegraphics[clip,width=0.8\columnwidth]{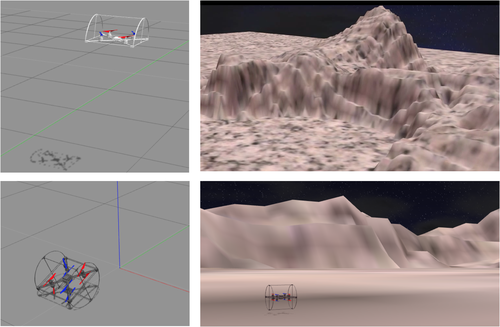}
        \caption{High-fidelity simulation based on ROS/Gazebo of the Shapeshifter on Titan. \textit{Top left:} simulated model of a Cobot. \textit{Top right:} model of the Sotra-Patera region, obtained by elevation maps of Titan. \textit{Bottom right:} the simulated Shapeshifter (assembled as Rollocopter) near Sotra-Patera on Titan. \textit{Bottom left:} the simulated model of the Shapeshifter assembled as Rollocopter.}
        \label{fig:gazebo_titan_0}
    \end{figure}
    
    \item \textbf{Shapeshifter's feasibility:} We compute first-order estimates of critical subsystems such as power, thermal, autonomy and communication for the Cobots and the Home-base. We determine that most subsystems, including power and thermal, can be implemented with current or high-readiness level technologies. In addition, we propose and experimentally validate the feasibility of a communication architecture based on mesh-networking. We highlight that future work should focus on detailing the design of the Cobots and of the Home-base from a thermal point of view, as well as detailing the mechanical design of the sample-collecting tool. In Phase I we focus on the feasibility of flight and ground operations, rather than under-liquid operations, which is left as future work.
    
    \item \textbf{Shapeshifter's autonomy:} We evaluate the feasibility of key algorithms necessary to achieve the full-autonomy, without the provision of human intervention. Due to the large communication delay between Earth and Titan, full autonomy is a necessary requirement for the Shapeshifter's mission. Our preliminary feasibility analysis is focused on collaborative mapping, collaborative localization, state estimation and task/motion planning for the Cobots. We propose a \ac{SLAM} framework that shows the feasibility of localizing in the different environments encountered on Titan, such as above-surface areas, canyons or caves. We additionally present different state estimation modalities (visual, thermal, laser, inertial) that allow the Cobots to navigate regardless of the presence of obscurants (methane fog, dust), and in vision-denied situations (during Titan's night, or in caves and cryolava tubes).
    
    \item \textbf{Case study of a mission to Titan's cryovolcano Sotra Patera:}  We show that the properties of the Shapeshifter allow the exploration of the possible cryovolcano Sotra Patera, Titan's Mare and canyons, satisfying the science requirements previously defined. We show that a Mothership similar to Cassini would be adequate to carry the Shapeshifter and the Home-base to Titan's orbit, and the similarity of the Home-base to the Huygens probe would make many aspects of the two missions (e.g. the EDL phase) similar. We define a ``baseline'' design for a Cobot and the Home-base for a mission to Titan's Sotra Patera region, showing that each Cobot would weight approximately $1.1$kg, including thermal control equipment and science payload, would use on average $31$W of power when in operation, and would have a volume of $0.03$m$^3$. Similarly, the Home-base would weight approximately $350$kg and would have a volume of $0.3$m$^3$. Each Cobot would carry a suction-based sample collecting tool, a stereo camera, an \ac{IMU} and other compact sensors such as a radiation monitor and a thermocouple. The Home-base, instead, would be the workhorse for scientific analysis and would be equipped with an x-ray spectrometer (based on the Mars 2020 \ac{PIXL}) and a chromatograph to study the sample of rocks and liquids collected by the Cobots. 
\end{itemize}

In summary, in the Phase I study we have detailed and studied some key-ideas and their implications for the Shapeshifter platform. We have critically analyzed some of the key aspects of the feasibility of the Shapeshifter, showing that our initial hypothesis of the feasibility of the platform is reasonable. The analysis performed during Phase I has underlined the importance of maturation and further investigation of some specific aspects, such as multi-agent autonomy (in terms of control for traversability of multiple terrains, localization in GPS denied environments), thermal control for compact flying robot and planning/decentralized coordination of multiple agents for mission operations. 
\par
The study draws on several existing and ongoing internal reports:
\begin{itemize}
    \item A. Tagliabue, S. Schneider, M. Pavone, A. Agha-mohammadi, ``The Shapeshifter:  a Multi-agent, Multi-modal mobility platform'', accepted for IEEE Aerospace Conference.
    \item A. Tagliabue, S. Schneider, M. Pavone, A. Agha-mohammadi, ``The Shapeshifter:  a Multi-agent, Multi-modal mobility platform'', in preparation for Acta Astronautica.
    \item S. Sabet, A. Agha-mohammadi, A. Tagliabue, S. D. Elliot, P. E. Nikravesh. ``Rollocopter: An Energy-aware Hybrid Aerial-ground Mobility for Extreme Terrains'', in 2019 IEEE Aerospace Conference 
    \item S. Schneider, A. Tagliabue, A. Agha-mohammadi, M. Pavone, ``Energy efficiency analysis of a multi-agent, flying and rolling platform'', Technical Report, Stanford University, October 2018.
    \item J. Baynord, G. Amos, J. Bennet, K. Bowers, A. Delaj, S. Khan, ``The Shapeshifter'', Technical Report, University at Buffalo: The State University of New York, December 2018.
    \item A. Agha-mohammadi, J. Hofgartner, P. Vyshnav, J. Mendez, D. Tikhomirov, F. Chavez, J. Lunine and I. Nesnas, ``Exploring Icy Worlds: Accessing the Subsurface Voids of Titan through Autonomous Collaborative Hybrid Robots'', International Planetary Probe Workshop (IPPW), Boulder, CO, June 2018. 
\end{itemize}

\counterwithin{figure}{chapter} %

\tableofcontents

\chapter{Introduction}
\begin{figure}
\centering
\includegraphics[width=0.8\columnwidth]{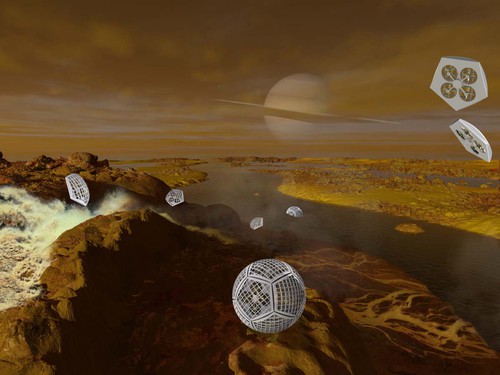}
      \caption{Artist's representation of the mission concept for the Shapeshifter exploring a liquid basin on Titan. The figure shows two Shapeshifters morphed in Rollocopters, and multiple Cobots flying nearby. Image credits: Ron Miller, Marilynn Flynn, Jose Mendez.}
\label{fig:intro}
\end{figure}
Titan is arguably the most Earth-like world in the Solar System, and its exploration has stimulated the interest of many scientists and researchers \cite{nasaNIAC}. It has a dense atmosphere, vast dune fields, rain, rivers, and seas that are part of a methane hydrologic cycle analogous to Earth’s water cycle \cite{aharonson2014titan}, \cite{hayes2016lakes}. Titan also has the most complex atmospheric chemistry in the Solar System, which may include prebiotic chemistry similar to that on early Earth before life began \cite{horst2017titan}. There is significant scientific motivation to explore this enigmatic world but two considerations have hampered mission concepts:
\begin{inparaenum}[(a)]
    \item many of the scientifically most enticing locations are difficult to access due to challenging surface conditions such as steep slopes, and 
    \item  a desire to explore all of Titan’s diverse terrains but the inability to traverse the long distances between them.
\end{inparaenum} \par Most of the current autonomous mobility systems for planetary exploration are based on monolithic (single agent) robotic platforms, and are usually tailored for the exploration of a specific terrain or domain, such as solid surface (e.g. landers \cite{lebreton2005overview}, rovers \cite{nasaMERS}, \cite{nasaMSL}, \cite{sauder2017automation} or hybrids \cite{pavone2013spacecraft}), liquid basins (e.g. \cite{stofan2013time}, \cite{oleson2015phase}), or atmosphere (e.g. \cite{lorenz2018dragonfly}, \cite{lorenz2008review}, \cite{barnes2012aviatr}, \cite{matthies2014titan}). In addition, because of their design philosophy, most platforms only provide redundancy at the component-level and thus do not guarantee margin for failure in high-risk, high-reward exploration scenarios. Platform designs specialized to traverse the challenging environments encountered in different bodies of the Solar System, such as rough surfaces \cite{elachi2005cassini}, \cite{bibring2005mars}, cliffs \cite{burr2013fluvial}, and subsurface voids \cite{tan2013titan}, incur penalties to the rest of the system design with only marginal mitigation of the risk. Unknown, unpredictable and constantly-changing \cite{hueso2006methane} environments additionally dictate for the need for flexible, redundant, multi-purpose mobility solutions, which are capable of efficient locomotion across domains.
\par In this report, we present a multi-agent robotic platform and mission architecture that allow for multi-domain, resilient, and all-access mobility on Titan - and potentially other bodies with an atmosphere. The proposed mobility system is based on the concept of shapeshifting, from which the name takes inspiration. The Shapeshifter is a collection of robotic units, called Cobots, capable of attaching and shaping themselves into different forms in order to accomplish various goals (see Figure \ref{fig:intro}). Each Cobot is mechanically simple and economically affordable, comparable to personal palm-size quadcopter, consisting of motor-propeller-based propulsion, instrumentation, and electronics. Via an active-controllable magnetic docking mechanism, multiple Cobots can mechanically connect together, morphing into new structures, which include 
\begin{inparaenum}[(a)]
\item a rolling vehicle that rolls on the surface, to increase the traverse range by reducing power consumption,
\item a flight array that can fly and hover above-surface and move in subsurface voids, like caves, and
\item a torpedo-like structure to swim under-liquid for chemical and biological measurements.
\end{inparaenum}
The proposed solution is complemented by a lander, called the Home-base, which hosts the power-source necessary to recharge the batteries of the Cobots, the scientific instrumentation and the telecommunication technologies. A swarm of Cobots can additionally be used to relocate the Home-base, using robust collaborative aerial transportation strategies \cite{collaborativeTransportation}, and to create a communication mesh capable of subterranean exploration and spatiotemporal measurements. 
Due to the inherent redundancy of the multi-agent platform and the low-cost design of each Cobot, high-risk tasks can be more easily accepted, as the loss of one or more agents does not compromise any of the functionalities of the platform. 
\par The proposed mobility solution is evaluated by showing the flying, docking and rolling capabilities of a simplified Shapeshifter prototype, constituted by two Cobots. In this simplified two-Cobot design, each Cobot is based on a traditional quadcopter design and is equipped with a custom-built hemicylindrical shell, adequate to fly, roll, and withstand small impacts. Additionally, our physical prototypes are capable of docking and un-docking via \ac{PM} based on \ac{PEM}, allowing the system to shape-shift into a cylindrical structure capable of rolling on different surfaces. By making use of first-principle laws, we derive a model to evaluate the energy efficiency of the pure rolling strategy and compare it with pure-flight strategies. We additionally present a high-fidelity simulation environment of Titan, used to test the proposed motion control algorithms and validate our energy-efficiency analysis. Our work is complemented by an analysis of the subsystems included in each Cobot and in the Home-base necessary to accomplish a described mission to Titan.%
\par The remainder of this work is structured as follows: Section \ref{sec:ScienceAndBusinessRationale} presents the science and business rationale for a mission to Titan and the possible cryovolcano Sotra Patera. Section \ref{sec:MobilityConcept} describes the proposed platform and details its mobility properties and mission capabilities. Section \ref{sec:RoboticPlatform} presents a simplified, two-agents prototype of a Shapeshifter and illustrates an energy-efficiency analysis and simulation. Section \ref{sec:Subsystems} describes some of the key subsystems required by each spacecraft. Section \ref{sec:Autonomy} presents some of the algorithms that can be employed to achieve the high levels of autonomy required by the platform. Finally, Section \ref{sec:MissionOperationsTitan} proposes a case study for the exploration of Titan, to establish some aspects of mission operations. Conclusion and future works are presented in Section \ref{sec:ConclusionAndFutureWorks}.

\chapter{Business case and science focus}
\label{sec:ScienceAndBusinessRationale}
Titan is a diverse and fascinating world with a wealth of exploration targets to expand our knowledge of the solar system. The moon of Saturn is mentioned more in the 2013 Decadal survey~\cite{board2012vision} than any other moon. It is the only place in the solar system, other than Earth, with a stable surface and radiation protection. Titan has an Earth-like meteorological process based on methane and a diversity of geological phenomena including cryovolcanoes like Sotra Patera~\cite{strobel2009atmospheric}. There is an abundance of organic synthesis that can give insight into the processes leading to life, and there may perhaps even be conditions to harbor life. These features of Titan make it a high priority exploration target where an immense amount can be learned in a variety of fields. Shapeshifter is designed to make the most of these diverse exploration interests through a mobile, multi-robot architecture that enables sampling and observations at many of the exciting areas of interest on Titan.  The multi-robot architecture also lends to a low-cost mission approach with simple Cobots and redundancy through multiple Cobots. 

\section{Science objectives of the exploration of Titan}
The science objectives that are targeted by Shapeshifter are outlined below, with reference to the three themes of the 2013 Decadal survey, before highlighting the game-changing elements brought by the Shapeshifter mission.

\subsection{The Shapeshifter and ``Building New Worlds" theme}
Titan, as a moon with gases and liquids, is an exploration target to understand how these gases came to the moon. Being too small to gravitationally attract the gases, there are questions on how the gases came to be; questions that give insight into the generation of the solar system. Are the gases from a solar nebula, or from a planetary subnebula? Are the gases all generated from within the moon with geological activity? As a lander, Shapeshifter can take in-situ measurements to inform analysis of the above hypotheses. These measurements include: abundances of volatiles, abundances of stable isotopes of oxygen, hydrogen and carbon, and abundances of noble gases. These measurements add to the body of knowledge with other bodies in the solar system to advance the current understanding of planetary and moon formation.

\subsection{The Shapeshifter and ``Planetary habitats" theme}
Titan is one of the most interesting places in the solar system to investigate the processes to generate conditions for life, and even the possibility of present life. The moon is often regarded as a laboratory for pre-biotic chemistry on a planetary scale with a rich array of organic genesis and complex organics~\cite{raulin2009titan}. For example, UV and plasma driven reactions in the nitrogen-rich upper-atmosphere generate complex organic molecules that condense into liquids as they fall on the surface. Understanding processes such as this is key to understanding the steps towards generating the chemicals that make up life. In-situ measurements both on the surface and in the atmosphere can provide insight into such processes, something described as ``One of the highest scientific priorities for the future" in the Decadal Survey. 
As part of studying these organic syntheses, there is particular interest on Titan in what types of complex organics are formed. Are there any amino acids or nucleotides, the currently understood building blocks of life~\cite{raulin2009titan}?
What Shapeshifter enables is measurements of these chemistries over a large range of Titan's diverse terrain: liquid/solid boundaries, lake sediment areas, dunes and planes as well as caves, faults and cryovolcanoes. Each of these areas may host different processes of generating organics, hence in exploring each of them, a more detailed picture of organic synthesis can be made. 

Titan not only provides an ideal laboratory to study the generation of organics, but it also has the potential to currently host conditions conducive to life. Titan has many of the what is currently understood to be required for water-based life: energy (solar), radiation protection, pressure (similar to Earth) and complex organics. If there was also water in places on Titan, such as might be possible with cryovolcanoes, then there could be all the ingredients for life. There are also theories of how methane-based life could exist, and in-situ measurements could test some of those theories by looking for the changes in concentration of H2 and reactive organics that could be from metabolizing lifeforms~\cite{mckay2005possibilities}. Some of the more interesting places for current life-supporting conditions are in difficult-to-reach places such as caves, volcanoes and lake boundaries: areas for which Shapeshifter is specifically suited. The processes and potential nature of life may well be different in each of these environments, making for a rich array of investigations that could be done with Shapeshifter.

\subsection{The Shapeshifter and ``Workings of Solar Systems" theme}
The diversity of geological features on Titan makes it a compelling exploration target to enhance our understanding of the geological processes throughout the solar system. Titan has nearly as much diversity as Earth, with planes, dunes, bedrock, mountain chains, lakes, rivers, and seas, as well as potentially: volcanic regions, faults, lake sediments and caves. Caves and cliff faces provide a particularly exciting target, with the ability to see layers of the geological history. The large range of possible features around the cryovolcano Sotra Patera, with cryo-lava caves, cliff faces and the volcano itself, make it a particularly interesting target for the Shapeshifter mission. This array of geological features lead to Titan being described as ranking ``near the top as a future target for exploration" in the Decadal, with the exploration of Titan have the potential for great advances in understanding the processes that shape and morph planets and moons~\cite{brown2009titan}. 

One of the strongest benefits of Shapeshifter is the ability to explore the many geological features that Titan holds. The mobile, flying sampling components of Shapeshifter (Cobots) allow many of these locations to be reached: lake edges, volcanoes, faults, caves and cliff faces. The redundant Cobots, in particular, make it possible to approach higher risk areas, such as cliff faces, that can give unparalleled insight into the geological history of the moon through the stratigraphy observable there. 

The exploration of Titan also provides insight into the different ways that climate can impact the conditions on a planet or moon, with a particular interest in the greenhouse and anti-greenhouse processes. As a surface based, mobile sensor system, Shapeshifter can allow measurements of the variations in temperature and pressure on the Titan, to supplement orbital measurements, for studying the variation in climate. The climate history could also be analyzed through the geological history observable from areas such as cliffs faces. In addition to climate, Shapeshifter could study meteorology, if an opportunity arose to make local measurements of a methane cloud, or storm, observations could be made from numerous locations at times.

\section{Unprecedented science opportunities}
Titan presents a thoroughly intriguing location for exploration, and one with a large diversity of phenomena. The architecture of Shapeshifter is designed to maximize the exploration of these phenomena, with the ability to have many agents (Cobots) with distributed sample collection - while having a well-equipped lab on the Home-base to do the required processing (e.g. a mass spectrometer). The particular science opportunities that are enabled by Shapeshifter include: 
\begin{itemize}
    \item Inspecting cliff faces to see the stratigraphy. Multiple, redundant flying robots make such a high-risk measurement feasible. These observations provide rich insight into geological and climate history. 
    \item Exploring caves. The Cobots can navigate rough and difficult terrain in fly or roll configurations to access underground environments, and can then work as a mesh communications network to get information out of the cave, back to the Home-base. In Section \ref{sec:Autonomy}, we will demonstrate some feasibility analysis and preliminary autonomy results in caves on physical flying robots.
    \item Exploring a large diversity of geological locations. The use of multiple Cobots, and their ability to efficiently traverse over rough terrain in flying or roll configurations, allows many geographically-disperse locations to be explored, including those with challenging terrain. 
\end{itemize}

\section{A design philosophy for new high-risk high-gain science opportunities at low cost}
The multi-agent nature of the Shapeshifter guarantees intrinsic system redundancy, unconventional in monolithic platforms such as traditional rovers and landers, where redundancy can be only guaranteed at a component level. If one or more Cobots become non-operational, for example, they can be easily replaced by other deployed units, as all the agents present the same design and capabilities. The total number of deployed Cobots can be large, as they are conceived to be economical to manufacture and deploy, thanks to:
\begin{itemize}
\item their mechanical simplicity, intrinsic in the multirotor design; 
\item their homogeneity, which makes the manufacturing cost low, as all the robots are the same;
\item their small size and volume, which significantly simplifies transportation and deployment.
\end{itemize} 
By design philosophy, the Cobots are also equipped with a minimal, low-cost scientific payload, mostly based on a sample collecting tool and imaging equipment, while the expensive sample analysis tools will be safely stored in the Home-base. 
\par Thanks to this platform architecture, it is possible to optimally trade-off the ability to accept a high-risk high-gain mission, by choosing the number of Cobots deployed, at low additional cost. The lander, which remains a critical part of the mission and constitutes the single point of failure, will be able to execute extremely low-risk operations thanks to the lack of mobility needs.

\section{Summary}
As one of the most compelling exploration targets in the Solar System, Titan holds many potential scientific findings. The variety of organic synthesis processes throughout Titan are particularly interesting, giving insights into the steps that lead to life. Titan also provides some of the richest diversity in geological features in the solar system, such as around the cryovolcano Sotra Patera and the Shapeshifter architecture grants the ability to explore these through multiple mobile flying Cobots. The design allows for high-risk, high-gain science to be attempted, such as exploring caves and inspecting cliff faces, while keeping costs constrained.

\chapter{Mobility concept and capabilities}
\label{sec:MobilityConcept}
\begin{figure}
\centering
\includegraphics[width=\columnwidth]{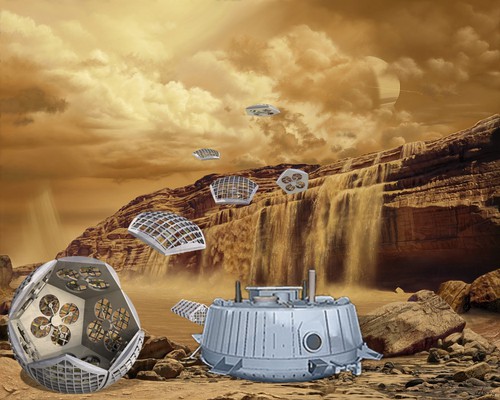}
      \caption{Artist's illustration of the Shapeshifter platform, constituted by Cobots (flying and partially self-assembled in a sphere, \textit{top and bottom left}) and by a lander called the Home-base \textit{(bottom right)}. Image credits: Ron Miller, Marilynn Flynn, Jose Mendez.}
\label{fig:PlatformOverview}
\end{figure}
In this section, we present the components that constitute the Shapeshifter platform \cite{shapeshifterIEEE}, namely the Cobots and the Home-base ground station, highlighting their primary functionalities. We additionally present the key locomotion modes that the Shapeshifter can adopt, leveraging the multi-agent nature of the platform, and the capabilities, in terms of science and mobility, that the Shapeshifter can offer. 

\section{Platform}
The hardware platform of the Shapeshifter is constituted by two components, the Cobots and the Home-base, as represented in Figure \ref{fig:PlatformOverview}.
\subsection{Cobot}
\begin{figure}[h]
\centering
\includegraphics[width=0.7\columnwidth]{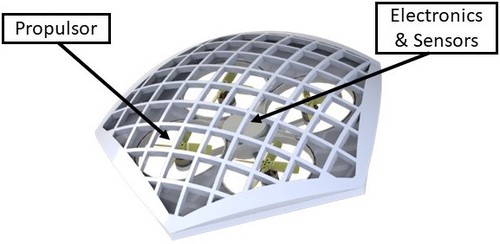}
      \caption{Artist's representation of a Cobot.}
\label{fig:Cobot}
\end{figure}
Shapeshifter’s Cobot \cite{aghacobotpatent} units are similar to existing, off-the-shelf quadcopters, as represented in Figure \ref{fig:Cobot}. Each Cobot is equipped with four rotating propellers that allow the Cobot to fly. The frame enclosing the Cobot's actuators is equipped with programmable polymagnets \cite{sullivan2005magnetic} that allow Cobots to self-assemble and perform shapeshifting. Cobot power is provided by an onboard battery that is recharged at the portable Home-base. Each Cobot is equipped with sensors to perform Visual-Inertial based navigation and mapping. Cobots will additionally be equipped with a tool to collect samples of rock and liquids to be analyzed by the Home-base. Their design is complemented by a radio, which allows communication with the Home-base and with other Cobots.

\subsection{Home-base}
The design of the Home-base is inspired by the Huygens lander used during the Cassini mission to Titan \cite{liechty2006cassini}.
The main task of the Home-base is to host:
\begin{inparaenum}[(a)]
  \item the instrumentation necessary to perform science measurements and analyze samples collected by the Cobots (such as a mass spectrometer),
  \item the radioisotope-based power system (RPS), such as a \ac{MMRTG} \cite{ritz2004multi} or a \ac{MSRG}, necessary to provide power to the Home-base itself and recharge the batteries of the Cobots, and
  \item to host the equipment necessary to establish a communication link with Earth.
\end{inparaenum}
The Home-base is not equipped with any mean of locomotion, but can be collaboratively transported by a swarm of Cobots (i.e. \cite{collaborativeTransportation}). An illustration of the Home-base, based on the Huygens lander design, is depicted in Figure \ref{fig:PlatformOverview}.

\section{Mobility modes}
\begin{figure}
\centering
\includegraphics[width=1\columnwidth]{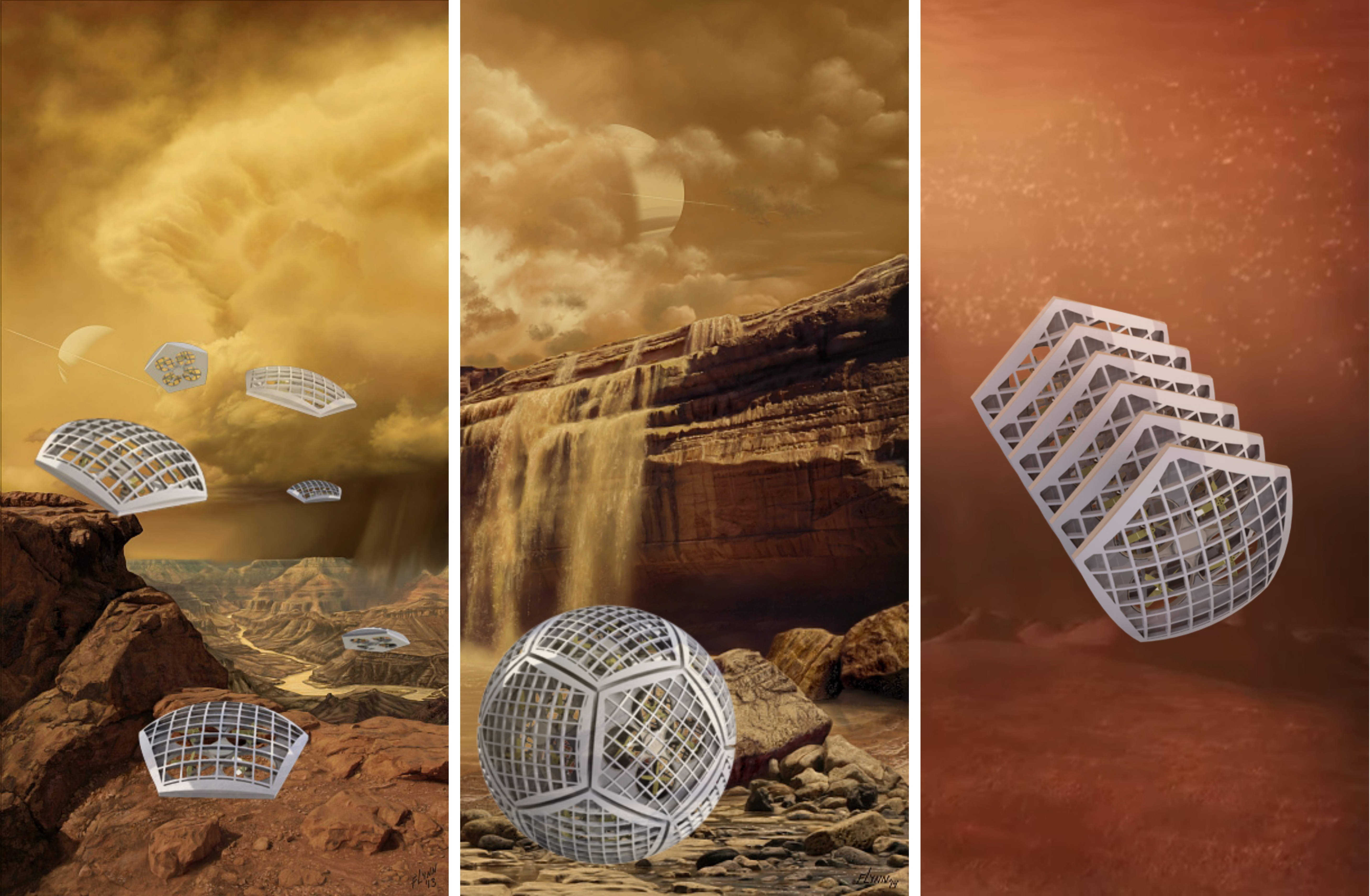}
      \caption{Artist's concept of the three main locomotion modes of the Shapeshifter (from left to right): flying, rolling on the surface and swimming. Background image credits: Ron Miller, Marilynn Flynn, Jose Mendez. }
\label{pic:sys_des:ref_frame}
\end{figure}
In this section, we present the main locomotion modes that the Shapeshifter can adapt by combining multiple Cobots together.

\subsection{Fly and Flying-array}
Each Cobot can autonomously fly in Titan's atmosphere, for exploration, mapping, and sample collecting purposes. A group of Cobots can additionally morph into a flight array, able to lift and carry heavy objects such as the portable Home-base.%

\subsection{Rollocopter}
Shapeshifter is able to morph into a spherical robot, the Rollocopter \cite{agharollopatent}, \cite{rollocopter2019Ieee}, that is able to roll on the surfaces and fly. In our artist's representation, we assume that 12 pentagon-shaped Cobots will morph into a spherical dodecahedron. While rolling, the robot is actuated by the same propellers used for flying, thanks to a locally centralized control strategy. The implications, in terms of control and energy efficiency, of rolling using the force produced by propellers are studied in our related work \cite{rollocopter2019Ieee}.
\subsection{Torpedo}
Each Cobot can autonomously swim under-water or float on the surface, relying on neutral buoyancy. Multiple Cobots can mechanically dock together in a torpedo-like structure for increased underwater autonomy or propulsion, in case of strong underwater currents. 

\label{sec:MissionArchitecture}

\begin{figure}
\centering
\includegraphics[width=1\columnwidth]{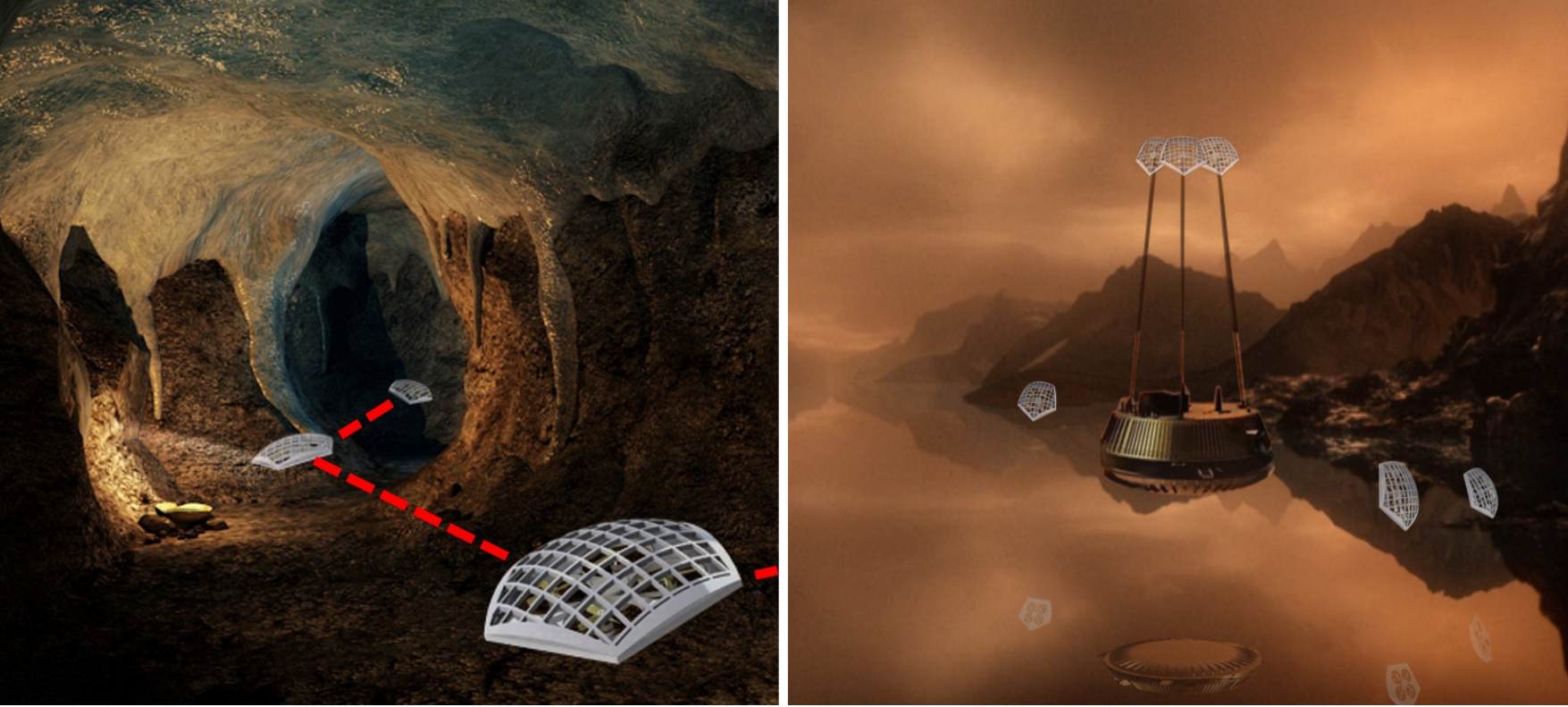}
      \caption{Artist's representation of (left) the Shapeshifter  exploring subterranean environments, maintaining a communication link with the surface, and (right) Shapeshifter carrying the Home-base using the Flying-array configuration (right). Image credits: Ron Miller, Marilynn Flynn, Jose Mendez.}
\label{fig:ExploringCaveAndCarryHomeBase}
\end{figure}

\section{Mission-relevant capabilities}
In this section, we present some of the key operational capabilities of the Shapeshifter platform. Such capabilities leverage the multi-agent nature of the system, as well as the different locomotion modes available and the possibilities of interaction with the Home-base; they constitute building blocks for the design of a mission on Titan using the Shapeshifter platform. The capabilities are grouped according to the three main locomotion configurations: Flying, Rollocopter and Torpedo. The estimates for the range and autonomy of the platform are based on the analysis presented in Section  \ref{sec:RoboticPlatform}.
\subsection{Flying and Flying-array}
\begin{itemize}
    \item \textbf{Mapping}: A group of flying Cobots can search for target science and create a detailed topographical map of Titan.  This allows the Shapeshifter to probe different kinds of surfaces, including very rough terrains and cliffs.
    According to the energy-efficiency analysis presented in Section \ref{sec:RoboticPlatform}, with a maximum range of 130 km, a team of Cobots flying radially outward from the Home-base could survey a circular area of over 10,000 km$^2$, and return home safely to recharge.
    \item \textbf{Comm-chain-array}: To maintain communication from kilometers under the surface (e.g., while exploring a cave), Shapeshifter can disassemble into Cobots that will maintain a network of communication nodes and outside line-of-sight from, e.g., a deep cryovolcanic lava tube to Titan’s surface. An artist's representation of this functionality is included in Figure \ref{fig:ExploringCaveAndCarryHomeBase}.
    \item \textbf{Collect spatiotemporal measurements}: By distributing multiple Cobots in a wide area, observations of a phenomenon can be correlated not only with respect to time, but also space. This can be useful to study spatiotemporal phenomena, e.g., the evolution of the storms present on Titan \cite{nasaTitanStorms}. 
    \item \textbf{Collect samples}: Cobots can collect samples of rocks and terrain using the onboard suction-based sample collecting tool. The samples are then analyzed by the Home-base. With approximately 30 N of maximum thrust available, a single Cobot could carry more than 20 kg of samples back to the Home-base. Maximum payload transportation capacity, anyway, could be reduced due to the aerodynamic drag of the payload (i.e. due to increased atmospheric density w.r.t. Earth), and collaborative strategies may be necessary. However, in this study, we are interested in very small samples, like dust and small rock fragments.
    \item \textbf{Transportation of the Home-base}: Shapeshifter can morph into a flight array of Cobots to lift and carry the portable Home-base from one mission site to another, as represented in an artist's concept in Figure \ref{fig:ExploringCaveAndCarryHomeBase}, by employing a decentralized collaborative aerial transportation strategy such as our related work \cite{collaborativeTransportation}. According to our initial estimates, $16$ Cobots will be enough to safely carry the Home-base, assuming a worst-case mass of $\approx 375$kg and taking into account controllability margins ($30\%$ of maximum thrust) of the Cobots. 
\end{itemize}
\subsection{Rollocopter}
\begin{itemize}
    \item \textbf{Traverse long distances}:  As shown in our analysis presented in Section \ref{sec:RoboticPlatform}, morphing into a sphere can be a more efficient way of traversing long distances. By taking advantage of energy-efficient mobility, the Shapeshifter can increase its range on a single charge to over 260 km, approximately doubling the range in flight mode. This corresponds to a reachable area of more than 50,000 km$^2$, radially outward from the stationary Home-base. %
    \item \textbf{\ac{Caver}}: Shapeshifter can explore subsurface voids, including cryovolcanic and karstic caves. For narrow passages, the Rollocopter configuration allows resilience to collisions. It allows the Cobots to bounce off of walls to go through cracks, holes, and narrow passages to reach science targets.
\end{itemize}
\subsection{Torpedo}
\begin{itemize}
    \item \textbf{Above and sub-surface navigation}: The propeller-based propulsion of each Cobot can generate thrust in gas and liquid environments \cite{diez2017unmanned}. Thus, Shapeshifter can navigate and explore above and below Titan’s mare, such as the Ligeia Mare represented in Figure  \ref{fig:LigeiaMare}. These functionalities can be further developed by leveraging studies on motors and propulsion systems for Titan’s hydrocarbon lakes \cite{hartwig2016exploring}.
\end{itemize}

\section{Summary}
In this section, we have presented the main morphing/mobility modes that the Shapeshifter platform offers, which enable the platform's navigation in multiple environments, such as liquids, subterranean or above the surface. By morphing into different shapes, the Shapeshifter can additionally provide unique mission-relevant capabilities, which range from creating a communication network to explore subterranean environments to transporting the Home-based or collect spatiotemporal measurements.

\chapter{Two-agents mobility concept analysis}
\label{sec:RoboticPlatform}
In this section, we present a simplified, two-Cobots prototype of the Shapeshifter. 
Despite featuring a reduced number of agents and a simplified mechanical design w.r.t. our original concept, this prototype shows critical feasibility aspects of the mobility of the platform,
such as the ability to create a mechanism for the docking and undocking of the two agents, and the ability to roll using the thrust produced by propellers. %
We additionally derive a dynamic model of the flying Cobot and rolling Shapeshifter, assuming no slipping. Such a model takes into account the power consumption, with special attention on the aerodynamic power, and it is used to compare the energy required to fly and roll on different terrains (in terms of friction and slopes) on Titan. We present a simple motion control algorithm developed to achieve the rolling behavior, which is validated, together with the energy-efficiency analysis, on a high-fidelity simulation environment of Titan. Last, we show preliminary experiments that validate our mechanical design and motion strategy, showing a docking, rolling and un-docking maneuver.
\section{Conceptual design for a simplified two-Cobot Shapeshifter prototype}

\begin{figure}
\centering
\includegraphics[width=1\columnwidth]{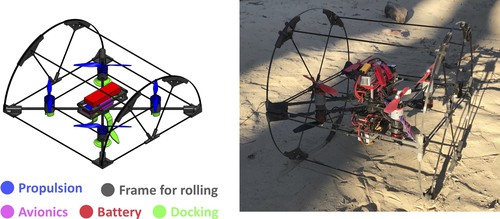}
      \caption{\textit{Left:} Sub-components of a Cobot prototype. \textit{Right:} Our physical prototypes of two Cobots} docked together forming a Rollocopter, capable of rolling on sand for increased range w.r.t. fly.
\label{fig:PlatformImplementationOverview}
\end{figure}

A mechanical design of the two-agents Shapeshifter is based on Cobots equipped with a hemispherical shell so that the docking of two agents creates a cylindrical structure adequate for rolling in one dimension. The prototype of the platform is represented in Figure \ref{fig:PlatformImplementationOverview}, where we have highlighted the different sub-components that constitute the Cobot. The two Cobots are identical, with the exception of the docking mechanism and the spin direction of the propellers. 

\section{Dynamic model}\label{sec:dynamics}
In this section, we describe the dynamic model of a flying Cobot and derive the dynamic model of a rolling Shapeshifter (Rollocopter), assuming that rolling happens without slipping. We additionally outline an induced-velocity model for power consumption as a function of the rotor thrust \cite{johnson2012helicopter}, both for the flying and rolling case. 

\begin{figure}
\centering
\includegraphics[width=1\columnwidth]{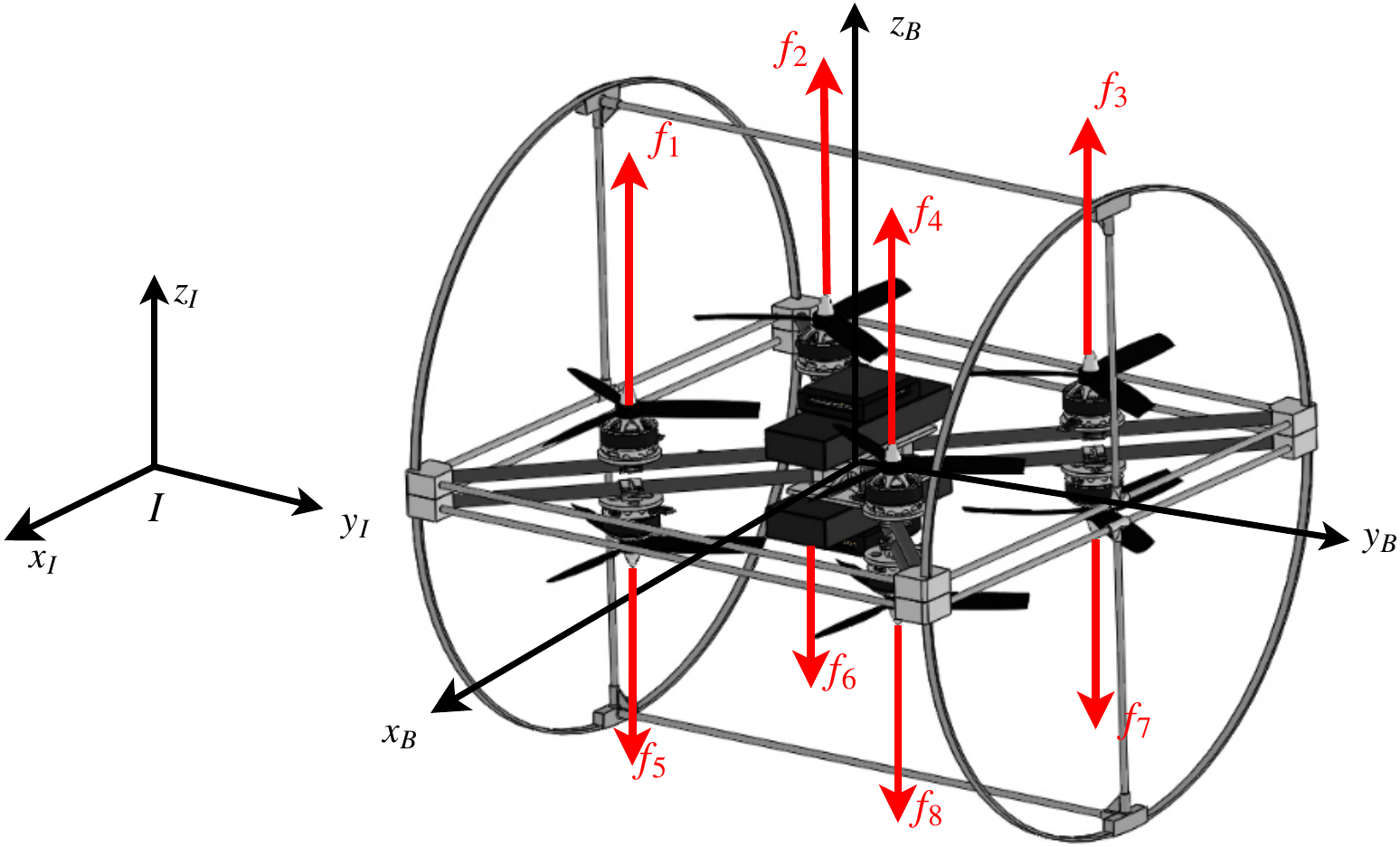}
      \caption{Model of the two-agents Shapeshifter, where we have highlighted the inertial reference frame $I=(x_I, y_I, z_I)$, the body-fixed reference frame $B=(x_B, y_B, z_B)$ and the forces produced by the actuators.}
\label{fig:3DShapeshifterFramesAndActuaturs}
\end{figure}

\paragraph{Reference frame}
For both the rolling and flying vehicles, we define an inertial reference frame $I=(x_I, y_I, z_I)$ a non-inertial reference frame $B=(x_B, y_B, z_B)$ fixed in the \ac{CoM} of each vehicle. The frames, in the case of the rolling robot, are represented in Figure \ref{fig:3DShapeshifterFramesAndActuaturs}.

\paragraph{Notation declaration:} $\prescript{}{W}{\boldsymbol{r}} = \prescript{}{W}{(r_x, r_y, r_z)}$ denotes a vector defined in the Cartesian reference frame $W$. The matrix $\prescript{}{A}{\boldsymbol{R}}_{BC}$ denotes the rotation matrix from the coordinate frame $C$ to $B$, defined in $A$.

\subsection{Rolling Shapeshifter}
The dynamic equations of the rolling Shapeshifter can be derived, for example, by using the Newton-Euler method, and are defined as: 

\begin{align}
& m\prescript{}{I}{\ddot{\boldsymbol{x}}} = \prescript{}{I}{\boldsymbol{R}}_{IB}\prescript{}{B}{\boldsymbol{f}_\text{cmd}} - m\prescript{}{I}{\boldsymbol{g}} + \prescript{}{I}{\boldsymbol{r}} + \prescript{}{I}{\boldsymbol{f}_\text{drag}} 
\label{eq:RollocopterDynamics:translation}
\\
&
\small \boldsymbol{J}\prescript{}{B}{\dot{\boldsymbol{\omega}}} = \prescript{}{B}{\boldsymbol{\tau}_\text{cmd}} - \prescript{}{B}{\boldsymbol{\omega}} \times \boldsymbol{J} \prescript{}{B}{\boldsymbol{\omega}} - l \prescript{}{I}{\boldsymbol{R}}_{IB}^{-1} (\prescript{}{I}{\boldsymbol{n}} \times \prescript{}{I}{\boldsymbol{r}}) + \prescript{}{B}{\boldsymbol{\tau}_\text{rolling}}
\label{eq:RollocopterDynamics:rotational}
\end{align}
where $m$ and $\boldsymbol{J}$ represent, respectively, the mass and inertia of the vehicle; the vectors $\boldsymbol{x}, \dot{\boldsymbol{x}}, \ddot{\boldsymbol{x}}$ represent the position of the robot and its derivatives, while $\boldsymbol{\omega}, \dot{\boldsymbol{\omega}} $ the angular rates and their derivatives. The attitude is represented by the rotation matrix $\prescript{}{I}{\boldsymbol{R}}_{IB}$, defining a rotation from $B$ to $I$, while $\prescript{}{I}{\boldsymbol{n}}$ and $\prescript{}{I}{\boldsymbol{r}}$ represent, respectively, the normal of the plane on which the vehicle is rolling and the reaction force with such a plane, expressed in $I$. The forces and torques produced by the actuators are defined by $\prescript{}{B}{\boldsymbol{f}}_\text{cmd} = (0, 0, f_\text{cmd})$, and $\prescript{}{B}{\boldsymbol{\tau}}_\text{cmd}$.  The total thrust force $f_\text{cmd}$ is defined as: 
\begin{align}
    f_\text{cmd} = \sum_{i=1}^{4}f_i - \sum_{i=5}^{8}f_i %
\end{align} 
under the assumption that all the propeller are placed on parallel planes. The propellers are placed so that opposite or adjacent pairs spin in opposite directions. For example, following the same numbering scheme adopted in Figure \ref{fig:3DShapeshifterFramesAndActuaturs}, the pair of propellers (1, 5) have opposite directions of rotation, as well as the pair (1, 2)). 
For simplicity, we assume that the vehicle rolls without slipping around $y_B$.
The torque due to the deformations caused by the interaction between the vehicle and the terrain 
is thus modeled as
\begin{align}
    \prescript{}{B}{\boldsymbol{\tau}}_\text{rolling} = (0, C_\text{rr} \prescript{}{I}{\boldsymbol{r}} \cdot \prescript{}{I}{\boldsymbol{n}} l, 0) 
    \label{eq:rolling}
\end{align}
where $\cdot$ represents the scalar product, $C_\text{rr}$ the rolling resistance coefficient and  $l$ the radius of the cylindrical shell of the robot.
\paragraph{Aerodynamic drag force}
We assume that the aerodynamic drag force $\boldsymbol{f}_\text{drag}$, applied in the \ac{CoM} of the robot, is function of the second power of the velocity of the vehicle:
\begin{align}
    \prescript{}{I}{\boldsymbol{f}}_\text{drag} = -\frac{1}{2}C_\text{d} \rho A ||\prescript{}{I}{\dot{\boldsymbol{x}}}||\prescript{}{I}{\dot{\boldsymbol{x}}}
    \label{eq:drag}
\end{align}
where $C_\text{d}$ is the drag coefficient, and $A$ is the aerodynamic area, computed as the area of a Cobot's rectangular base projected on the plane orthogonal to the velocity of the vehicle $\prescript{}{I}{\dot{\boldsymbol{x}}}$. We observe that scalar $A$ is a function of the attitude of the robot.

\subsection{Flying Cobot}
The dynamic model of a flying Cobot can be obtained by Equations (\ref{eq:RollocopterDynamics:translation}) and (\ref{eq:RollocopterDynamics:rotational}) assuming $\boldsymbol{r} = \boldsymbol{0}$ and, as a consequence, $\boldsymbol{\tau}_\text{rolling} = \boldsymbol{0}$. Additionally, it is worth noting that every Cobot has only four propellers, all pointing on the same direction (e.g. $f_1, ..., f_4$).

\subsection{Power consumption model for rolling and flying robots}
We assume that the total power consumption of the robot is directly proportional to the aerodynamic power produced by the propellers, according to the commanded thrust and the motion of the robot, similar to \cite{ware2016analysis}. We disregard the power consumption of other processes (e.g. thermal) in our analysis, as we don't expect many discrepancies between the two mobility modes.
Our induced velocity model relates individual rotor thrust to aerodynamic power, as detailed in \cite{leishman2006principles}, assuming forward-flight regime. %
The forward flight power-thrust model for the $i^\text{th}$ propeller can thus be expressed as:
\begin{align}
    P_i = \frac{f_i(\nu_i - \nu_\infty \sin \alpha)}{\eta_p \eta_m \eta_c}
\end{align}

Power consumption of rotor $i$, $P_i$, is defined as a function of rotor thrust ($f_i$), induced velocity ($\nu_i$), freestream airspeed ($v_\infty$), and angle of attack ($\alpha$). For this analysis, we assumed a propeller efficiency of $\eta_p$ = 0.6, motor efficiency of $\eta_m$ = 0.85, and controller efficiency $\eta_c$ = 0.95. 
We assume that this forward-flight model applies to the rolling configuration as well, justified by the observation that all rotors in the Rollocopter are moving forward relative to freestream at all times during a rotation. We neglect any effect that potential vortex ring states might have on total power consumption.

\section{Energy efficiency analysis}

In this part, we introduce the approach used for the energy analysis \cite{shapeshifterStanford}.
We focus on the steady-state, planar motion of the robots ($\ddot{\boldsymbol{x}} = \boldsymbol{0},  \dot{\boldsymbol{\omega}} = \boldsymbol{0}$), and assume that the motors are controlled to apply pure torque (zero net thrust), implementing the presented control strategy. In the following analyses, $\theta$ defines the slope of the terrain, and $\alpha$ defines the angle between the body frame and the inertial coordinate system, as rotated about $\hat{y}_I = \hat{y}_B$ (\textit{pitch} angle).

\subsection{Rolling Shapeshifter}
\begin{figure}[htp]
    \centering
    \includegraphics[width=0.8\linewidth]{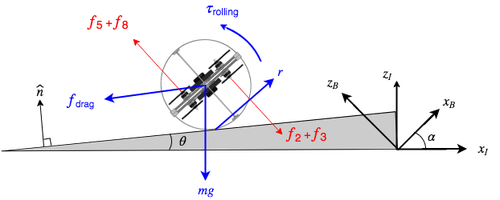}
    \caption{Free body diagram: rolling uphill. Thrust from four rotors produces a pure torque. Rollocopter is propelled by the ground reaction force (assume rolling without slipping) up a slope defined by $\theta$.}
    \label{fig:rolling_fbd}
\end{figure}

The free-body diagram used in the planar-motion analysis of the rolling configuration is shown in Fig. \ref{fig:rolling_fbd}. Torque due to rolling resistance is calculated as per Eq. (\ref{eq:rolling}). This analysis assumes a rolling friction coefficient between that of consolidated soil ($C_\text{rr} = 0.01$) and tires on loose sand ($C_\text{rr} = 0.2$), consistent with data gathered from the Huygens Surface Science package \cite{zarnecki2005soft}.
Drag force on the Rollocopter is calculated based on Eq. (\ref{eq:drag}), where drag coefficient, $C_\text{d} = 2.1$, is approximated based on Titan's atmosphere \cite{liechty2006cassini}. The aerodynamic area, $A$, is computed as the area of a Cobot's rectangular base projected onto the plane orthogonal to its velocity vector, as a function of the Rollocopter's orientation (\ref{eq:Ax}). Cobot dimensions and other parameters used for the analysis are defined in Table \ref{table:dimensions}.

\newcommand\Tstrut{\rule{0pt}{2.6ex}}       %
\newcommand\Bstrut{\rule[-0.9ex]{0pt}{0pt}} %
\newcommand{\TBstrut}{\Tstrut\Bstrut} %

\begin{table}[h]
\begin{center}
 \begin{tabular}{| r l c |} 
 \hline
 \textit{Symbol} &  \textit{Meaning} & \textit{Value} \TBstrut\\
 \hline \hline
 $h$ & \textit{rolling}: height from one rotor & 0.16 m \TBstrut\\
  & to opposite rotor & \TBstrut\\
  & \textit{flying}: height from rotor to base & 0.08 m \TBstrut\\
 \hline 
 $l$ & radius of cylinder & 0.2 m \TBstrut\\
 \hline
 $w$ & width of cylinder & 0.4 m \TBstrut\\
 \hline
 $C_\text{d}$ & drag coefficient & 2.1 \TBstrut\\
 \hline 
 $C_\text{rr}$ & rolling resistance coefficient & 0.01 - 0.2 \TBstrut\\
 \hline 
\end{tabular}
\end{center}
\caption{Relevant model parameters for the two-Cobots Shapeshifter, used for the energy-efficiency analysis.}
\label{table:dimensions}
\end{table}

\begin{equation}
    A = (h \mid\cos\alpha\mid + 2l\mid\sin\alpha\mid)w
    \label{eq:Ax}
\end{equation}

\subsection{Flying Cobot}
\begin{figure}[h]
    \centering
    \includegraphics[width=0.6\linewidth]{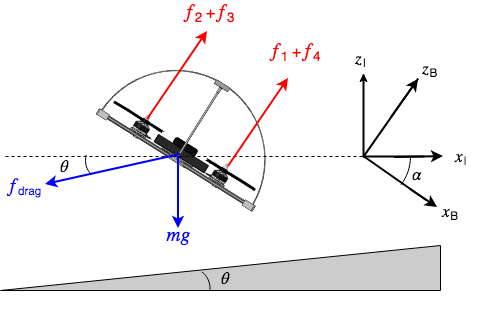}
    \caption{Free body diagram: flying. Cobot flies at a constant height above a hill defined by $\theta$. Angle of attack $\alpha$ is determined by the angle required to keep the Cobot at a constant altitude.}
    \label{fig:flying_fbd}
\end{figure}

Fig. \ref{fig:flying_fbd} shows a free-body diagram for the flying configuration of a single Cobot. In this configuration, $\alpha$ is determined based on the angle of attack required for the Cobot to fly at a constant height above the ground. Aerodynamic drag force is again given by (\ref{eq:drag}), and aerodynamic area by (\ref{eq:Ax}).

\section{Motion control}
\subsection{Rolling Shapeshifter}
\label{sec:rolling_control} 
For this initial feasibility study, we implemented a simple, centralized control strategy that tracks a desired body rate $\prescript{}{B}{\boldsymbol{\omega}}_\text{des}$ by applying a pure torque on the Rollocopter, i.e., producing zero net thrust. A detailed derivation for the control strategy of a conceptually similar platform is proposed in our related work \cite{rollocopter2019Ieee}. The desired torques $\prescript{}{B}{\boldsymbol{\tau}}_\text{cmd}$, expressed in the body-fixed reference frame $B$ of the rolling vehicle, are computed according to a proportional-integral (PI) controller, using the measurements $\prescript{}{B}{\hat{\boldsymbol{\omega}}}$ of the body angular rates provided by the IMU on-board one of the two docked Cobots:
\begin{align}
    &\boldsymbol{e} = \prescript{}{B}{\boldsymbol{\omega}}_\text{des} - \prescript{}{B}{\hat{\boldsymbol{\omega}}} \\
    \prescript{}{B}{\boldsymbol{\tau}}_\text{cmd} & = \boldsymbol{K}_\text{p}\boldsymbol{e} + \boldsymbol{K}_\text{i}\int_{t_0}^{t} \boldsymbol{e} dt
\end{align}
where $\boldsymbol{K}_\text{p}$ and $\boldsymbol{K}_\text{i}$ are diagonal matrices, tuning parameters of the controller. 
\subsubsection{Torque allocation strategy}
Given the commanded torque $\prescript{}{B}{\boldsymbol{\tau}}_\text{cmd}$, a rotation speed input $n_i$ for the $i\textit{-th}$ propeller, with $i = 1, ..., 8$ is produced in the following way.
First, we define the quadruple $(f_\text{A}, ..., f_\text{D})$, where every entry corresponds to the force produced by each pair of opposite propellers (e.g. $f_\text{A} = f_\text{1} - f_\text{5}$, $f_\text{B} = f_\text{2} - f_\text{6}$, etc...).
Second, we derive an expression which relates the forces $f_\text{A}, ..., f_\text{D}$ with
\begin{inparaenum}[(a)]
 \item the total torques produced by the propellers, which we assume to be equivalent to $\prescript{}{B}{\boldsymbol{\tau}}_\text{cmd}$, and
 \item the force $f_\text{cmd}$ produced along the positive $z$ axis of $B$, and expressed in body $B$ frame.
\end{inparaenum} 
Such expression is defined as:
\begin{equation}
    \begin{bmatrix}
    f_\text{cmd} \\ \boldsymbol{\tau}_\text{cmd} \\
    \end{bmatrix} =
    \boldsymbol{M} 
    \begin{bmatrix}
        f_\text{A} \\ f_\text{B} \\ f_\text{C} \\ f_\text{D} \\
    \end{bmatrix}
    \label{eq:AllocationStrategy}
\end{equation}
where $\boldsymbol{M}$ is a squared, full-rank matrix by the design of the system, and corresponds to the inverse of the allocation matrix. The matrix $\boldsymbol{M}$ is defined as: 
\begin{align} 
\boldsymbol{M} = 
    \begin{bmatrix}
    1 & 1 & 1 & 1 \\
    -c & c & c & -c \\
    -c & -c & c & c \\
    -k_\tau & k_\tau & -k_\tau & k_\tau \\
    \end{bmatrix}
\end{align}
with $c = \frac{a}{\sqrt{2}}$, where $a$ is the arm length of each propeller to the \ac{CoM}, and $k_\tau$ is a constant with maps the $i- \textit{th}$ propeller's thrust $f_i$ to the aerodynamic torque $\tau_i$, according to $\tau_i = k_\tau f_i$.

Last, because we have assumed that the net force produced by propellers has to be zero, we impose $f_\text{cmd} = 0$ and solve Equation \ref{eq:AllocationStrategy} by inverting the square matrix $\boldsymbol{M}$, 

allowing us to find the desired force that has to be produced by each propeller pair.
The desired rotation speed to be commanded to the propellers in each propeller pair can be retrieved in a straightforward way, for example by checking, in the case of the propeller pair A (constituted by propeller 1 and 5):
\begin{equation}
    (n_1, n_5) 
    = 
    \begin{cases}
        (\sqrt{f_\text{A}/k_\text{t}}, 0) & \mbox{if } f_\text{A} \geq 0 \\
        (0, \sqrt{-f_\text{A}/k_\text{t}}) & \mbox{if } f_\text{A} < 0 
    \end{cases}
\end{equation}
where $k_\text{t}$ relates the force produced by each propeller with its rotation speed according to $f_i = k_\text{t} n_i^2$.

\subsection{Flying Cobot}
For the flying Cobot, we employ a common cascaded control architecture \cite{Controll67:online} provided by the embedded microcontroller framework \cite{meier2015px4}.

\section{Prototype and design considerations}
\begin{figure}
    \centering
    \includegraphics[width=0.9\textwidth]{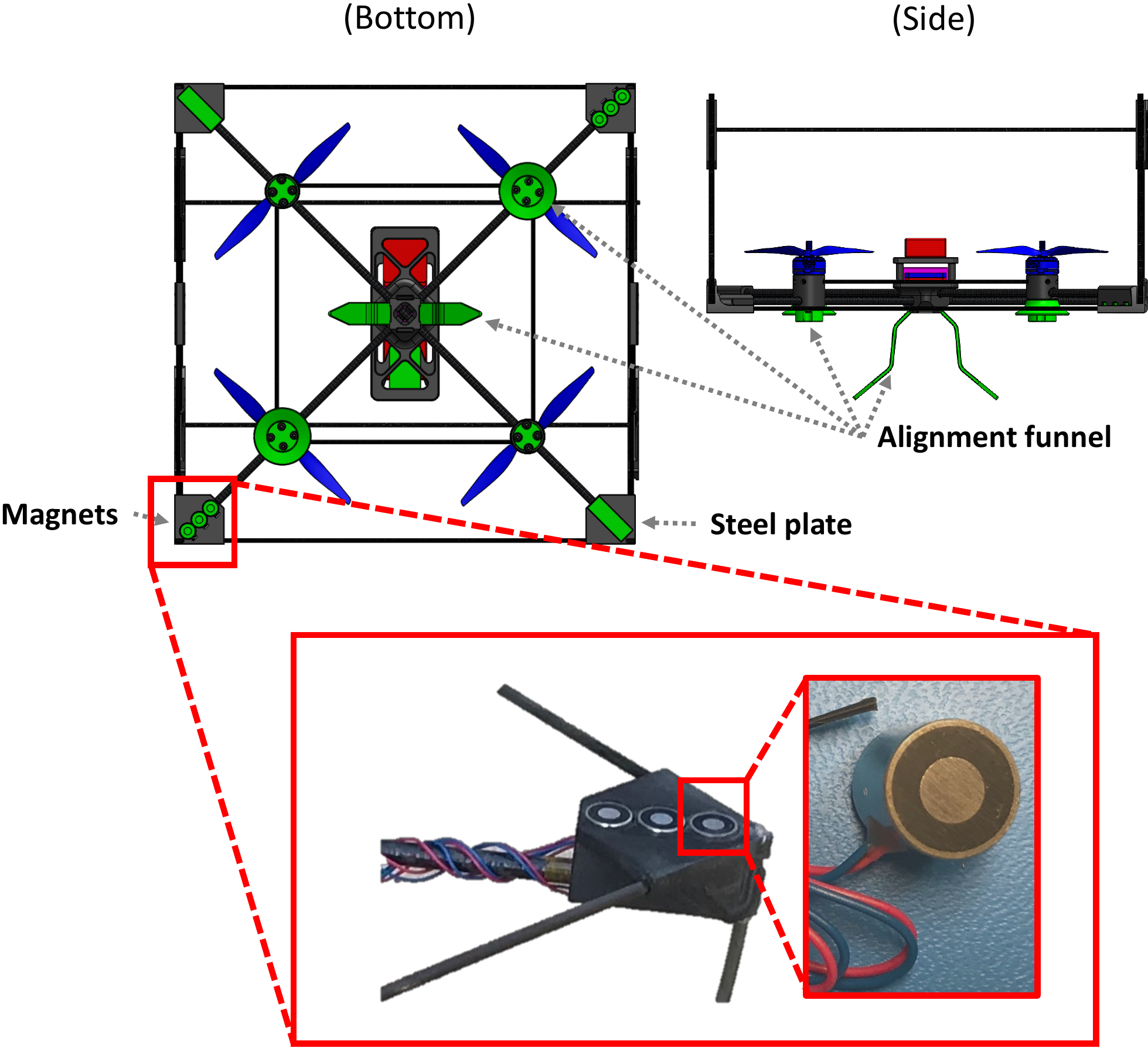}
      \caption{Illustration of the magnetic and mechanical docking mechanism used to connect two Cobots.}
    \label{fig:DockingMechanism}
\end{figure}
Each Cobot consists of four 6-inch propellers, \textit{EMax 2300 kV} brushless-DC motors, and a three-cell 2200 mAh battery, which provides approximately eight minutes of flight (on hover, on Earth). The side length and the diameter of the cylinder when two Cobots are docked is 0.4 m; the total weight of each Cobot is approximately 0.8 kg. Enough payload transportation capacity is guaranteed by the maximum thrust produced by the propellers, which corresponds to approximately 32 N. On-board computing power and IMU are provided by a Pixwhawk-mini running a PX4 flight stack \cite{meier2015px4}. The two Cobots are identical, with the exception of the position of the magnets and mechanical funnels for the docking mechanism.
\paragraph{Shell for rolling and flying: }
Each Cobot is equipped with a shell adequate for flying and capable of withstanding small impacts during rolling. The shell is designed using lightweight carbon fiber tubes to satisfy weight and stiffness requirements for this task. The carbon fiber tubes are connected together via 3D printed joints.
\paragraph{Docking mechanism:}
The docking mechanism is based on 12 permanent electromagnets (PEM) mounted at two diametrically opposite extremities of each Cobot. Each PEM weights approximately 10 g and produces a normal force of 15 N when connected to a 2 mm thick plate of steel, and when not powered. When powered, the magnet produces approximately 0 N of force. Based on this property, no power is required to maintain the two agents docked. The PEM can be activated and deactivated by a microcontroller interfaced with the onboard computer.
The docking mechanism is additionally constituted by mechanical funnels connected at the bottom of each Cobots, used to compensate for misalignments during the docking phase and to cancel the shear forces between the agents, during rolling. A representation of the docking mechanism, where its main components have been highlighted, can be found in Figure \ref{fig:DockingMechanism}.

\section{Simulation Environment}
One goal of this work is to construct a physics-based simulation to verify the analytical energy analysis, as well as provide a tool for further mobility analyses and development of control strategies. We chose to use Gazebo \cite{koenig2004design} as the simulation environment because of its realistic physics engine, including aerodynamic drag and rolling resistance, as well as existing quadrotor packages \cite{Furrer2016}. With Gazebo, the user can place Cobots in either flying or two-agents rolling configurations, then input waypoints for the agent; the simulation outputs power and energy data in real-time, providing a versatile platform for testing different Shapeshifter missions.
\par
\begin{figure}
    \centering
    \includegraphics[clip,width=\textwidth]{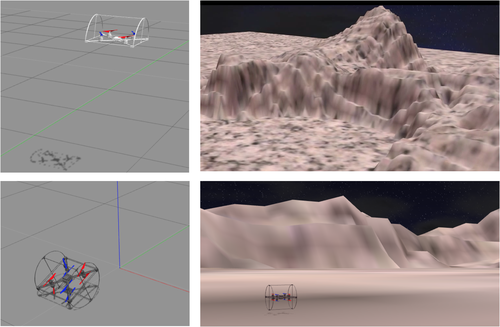}
    \caption{High-fidelity simulation based on ROS/Gazebo of the Shapeshifter on Titan. \textit{Top left:} simulated model of a Cobot. \textit{Top right:} model of the Sotra-Patera region, obtained by elevation maps of Titan reconstructed from images captured by the Cassini mission. \textit{Bottom right:} the simulated Shapeshifter (assembled as Rollocopter) near Sotra-Patera on Titan. \textit{Bottom left:} the simulated model of the Shapeshifter assembled as Rollocopter.}
    \label{fig:gazebo_titan}
\end{figure}

 Our simulation setup is not limited to simulate the robot, but includes a 3D model of the Sotra Patera area on Titan. This model has been obtained via an elevation map computed from images captured by the Cassini mission.
Fig. \ref{fig:gazebo_titan} \textit{(bottom)} shows the Shapeshifter rolling on a surface generated from a depth elevation map of Sotra Patera on Titan. We also consider a basic simulation with one flying Cobot and one Rollocopter traversing a flat surface, shown in Fig. \ref{fig:gazebo_titan} \textit{(top)}, to isolate the mobility primitives.
\par 
In Phase II, we plan to leverage the advanced multi-body dynamics simulator DARTS \cite{NASAJetP19:online} developed at JPL in order to more accurately  model the terramechanic interactions between Titan and the Shapeshifter during surface mission operations.

\section{Experimental and simulation results}
In this section, we present the preliminary results for the mobility validation and energy-efficiency analysis of the two-agents Shapeshifter \cite{video}. The results highlight the mechanical feasibility of having a platform capable of both flying and rolling on sandy terrain, as well as the energy savings realized by morphing into a rolling vehicle. In this section, we additionally introduce a high-fidelity simulation environment of Titan, developed to test our algorithms, validate our analysis and provide a way to simulate mobility aspects of the proposed mission.

\subsection{Docking, rolling and un-docking experiments}
\begin{figure}
    \centering
    \includegraphics[width=\linewidth]{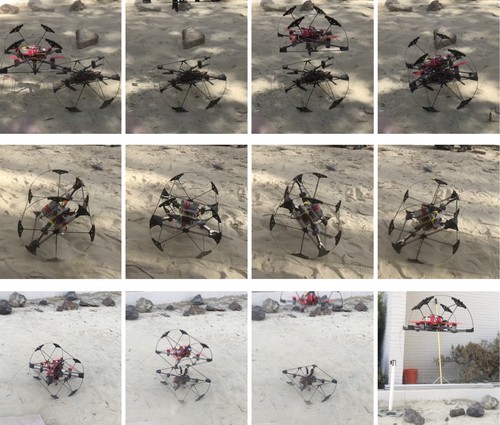}
    \caption{Frames from the video clip of the experiments conducted with our two-Cobot Shapeshifter.\textit{Top:} Docking sequence. \textit{Center:} Rolling sequence. \textit{Bottom:} Un-docking sequence.}
    \label{fig:shapeshifter_movie}
\end{figure}
In this part, we present the experimental results of the docking, un-docking and rolling maneuver obtained with our prototype, which are represented in Figure \ref{fig:shapeshifter_movie}. 
\paragraph{Docking} As represented in the first row of Figure \ref{fig:shapeshifter_movie}, we fly the first Cobot while the second agent is on the ground, with the propellers pointed towards the ground. This experiment shows that docking is possible despite the limited flying accuracy of a human pilot, thanks to the alignment funnels and the strong magnetic field created by the magnets. We verify that docking has successfully happened by manually lifting the top Cobot (after docking) and observe that the entire Shapeshifter (both Cobots) are being lifted.
\paragraph{Rolling} Once docked, all the motors of the agents are manually connected to the onboard controller of one robot. This limitation will be overcome in future works, for example by establishing a wireless link between the Cobots or by developing a decentralized control strategy. In our experiment, shown in the second row of Figure \ref{fig:shapeshifter_movie}, a pilot controls the rolling motion via remote control (RC), configured to send the desired angular rates $\boldsymbol{\omega}_\text{cmd}$. Preliminary experiments show that the vehicle can effectively roll on sandy terrain, uphill and in small dunes. The cylindrical design, anyway, severely limits the yawing (turning) capabilities of the robot. \paragraph{Un-docking} The \acp{PEM} have been configured so that they can be remotely turned off. The experiment is shown in the third row of Figure \ref{fig:shapeshifter_movie}.

\subsection{Energy-efficiency analysis on Titan via simulations}
\begin{figure}
    \centering
    \includegraphics[width=0.8\linewidth]{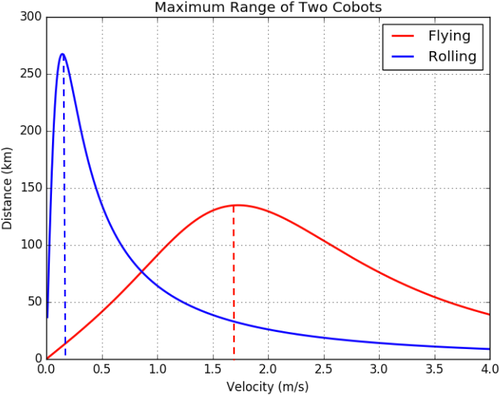}
    \caption{Range vs. velocity for two Cobots on Titan, assuming that the power consumption of the robot is directly proportional to the power used by the actuators. The maximum achievable range for flying and rolling along a surface of consolidated soil. Dashed lines indicate optimal velocity for each configuration.}
    \label{fig:range_v_velocity}
\end{figure}

\begin{figure}
    \centering
    \includegraphics[width=0.82\linewidth]{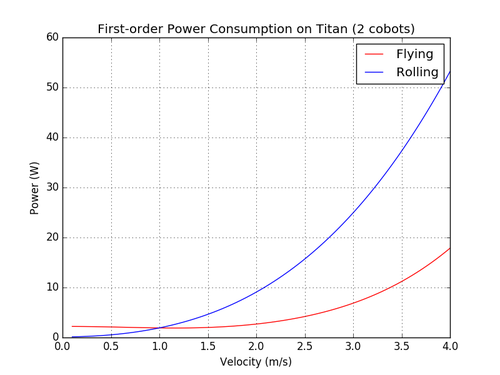}
    \caption{Power vs. velocity for two Cobots on Titan. }
    \label{fig:power_v_velocity_titan}
\end{figure}

\begin{figure}
    \centering
    \includegraphics[width=0.7\linewidth]{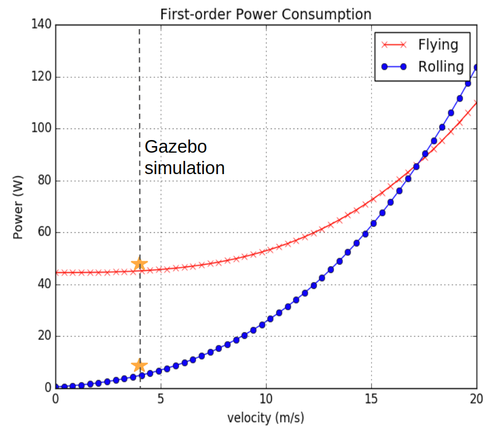}
    \caption{Power vs. velocity for two Cobots on Earth. }
    \label{fig:power_v_velocity_earth}
\end{figure}

In this section, we aim to use our dynamic model to determine environmental conditions for which rolling is the more energy-efficient mobility primitive, as well as the conditions for which it is more efficient to fly. To make this distinction, we focus on developing a functional relationship between terrain primitives and the required energy of mobility. Since the proposed mission involves traversal over long distances of Titan's surface, the main objective of this analysis is to determine the maximum expected steady-state range of the Shapeshifter, both for rolling and flying. 
The following results employ the analytical model and pure torque control strategy, 
applying physical parameters for Titan: $g$ = 1.352 m/s$^2$ and $\rho$ = 5.4 kg/m$^3$. To get numerical results, we assume each Cobot has a 870 kJ battery (2200mAh at 11V).
Fig. \ref{fig:range_v_velocity} shows the maximum achievable range for both flying and rolling along a surface of consolidated soil ($C_{rr} = 0.01$); the dashed line indicates the velocity that maximizes range for that configuration. At an optimal velocity of 0.14 m/s, two rolling Cobots are able to travel 267 km, while the maximum range for two flying Cobots is only 135 km, traveling at their optimal velocity of 1.7 m/s. Assuming that we operate at the optimal range speed, the power required by the actuators of each Cobot to roll is thus approximately $1$W, while the power required to fly is approximately $10$W. In Figure \ref{fig:power_v_velocity_titan}, we additionally show the total power consumption for the two-agents Shapeshifter on Titan, while on Figure \ref{fig:power_v_velocity_earth} we illustrate the comparison on Earth. We can observe that the total power required to roll or fly at the optimal velocity on Titan is low if compared to the nominal power consumption of an equivalent quadcopter on Earth, showing the opportunity of long-range missions on Titan.


\begin{figure}
    \centering
    \includegraphics[width=0.8\linewidth]{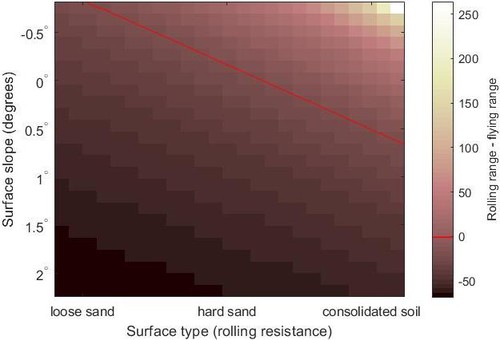}
    \caption{Advantage of rolling vs. flying, as it relates to the surface slope and rolling resistance. Flying range is on the order of 130km for all terrains; add flying range to results shown here to get total rolling range.}
    \label{fig:multi_dim}
\end{figure}

As expected, rolling is more efficient than flying over ideal surface conditions. To optimize energy usage while traversing a non-ideal region, we must develop a relationship between terrain characteristics and power required for traversal. Fig. \ref{fig:multi_dim} considers terrains that vary in surface traction from consolidated soil to loose sand, and in steepness from -0.5$^\circ$ to +2$^\circ$. For each surface type and mobility method, we compute maximum range assuming the agents travel at their corresponding optimal velocities. By considering the difference in achievable range for each configuration, we can see where rolling is favorable (above the red line) vs. flying (below line).
These results demonstrate that neither rolling nor flying consistently outperforms the other; rather, each configuration optimizes energy efficiency for a different set of conditions.
Especially since the characteristics of Titan's surface are largely unknown, a shape-shifting platform is crucial to accommodate unexpected surface conditions.
Furthermore, this relationship between terrain and mobility allows us to build a traversability map, and ultimately plan energy-efficient trajectories that optimize the Shapeshifter's route as well as mobility configuration, to take full advantage of the multi-modal architecture.

\subsection{Energy-efficiency analysis for more than two Cobots}
\begin{figure}
    \centering
    \includegraphics[width=0.7\linewidth]{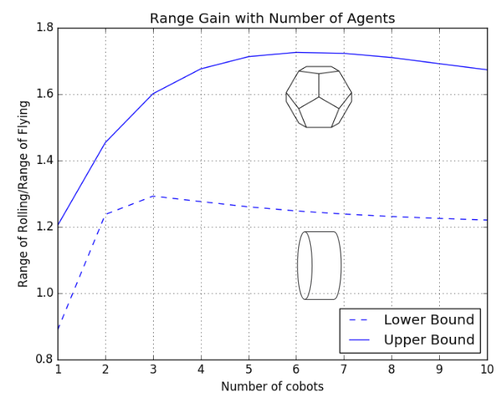}
    \caption{Advantage of rolling vs. flying, as a function of the number of agents. The range is shown as a ratio of maximum rolling range to maximum flying range.}
    \label{fig:range_v_num_agents}
\end{figure}
To explore more facets in the Shapeshifter's potential efficiency improvements, we considered how the advantage of rolling over flying might change by incorporating more Cobots into the rolling configuration. Specifically, since the cost of flying a certain distance scales linearly with the number of agents, and the cost of rolling is hypothesized to scale at a lower rate, rolling more than two agents together would make rolling increasingly advantageous. This analysis serves to critically examine the relationship between the maximum range of the platform when rolling and the number of Cobots that compose each Shapeshifter, assuming approximately constant the size of each Cobot. For this study, we explicitly consider the drag of the cage (in previous analyses, cage drag was neglected), isolating the worst-case aerodynamics of each cage design.
\par
Fig. \ref{fig:range_v_num_agents} presents a lower and upper bound for the advantage of rolling on a flat surface (quantified as the ratio of maximum rolling range to the maximum flying range). To calculate the upper bound for rolling distance, we consider $n$ cobots (each modeled as a rectangular prism the same size and mass as in the 2-cobot case), arranged in a ``pseudo-platonic solid'' encapsulated by a spherical shell. Given $n$ cobots, we use linear interpolation of real platonic solid sizes (4, 6, 8, or 12 faces) to determine the spherical radius that could encapsulate a ``pseudo-platonic solid'' with $n$ faces. That sphere's circular cross-sectional area is then used in the computation of aerodynamic drag. To determine the lower range bound, we simply expand a cylinder, similar to the one we have been considering, to include more Cobots. The cross-section of an $n$-agent Shapeshifter in this configuration would look like a regular $n$-sided polygon. This lower-bound case results in a rectangular cross-sectional area that grows quickly with $n$. In both the lower and upper bounding cases, we still assume that thrust applies a pure torque with a net zero force.
The resulting bounded area emphasizes the importance of aerodynamic efficiency and Cobot configurations. We notice an unambiguous improvement from one to two agents, suggesting that the multi-agent approach provides a more efficient system as compared to an individual agent rolling in a single attached cage. The results also suggest that even in the worst case, we see a marked advantage rolling as a cooperative multi-agent robot, rather than flying independently.

\section{Summary}
In this chapter, we have shown the mechanical and algorithmic feasibility of building a multi-agent platform, whose agents are able to fly, dock and un-dock using magnets by building and designing the motion control algorithms for a  two-agents prototype of the Shapeshifter. We have additionally performed an energy-efficiency analysis, comparing the power cost of flying and rolling, showing that rolling can be up to two times more energy efficient than flying for a two-agents Shapeshifter. We compute a first order estimate (only based on the power consumption of the actuators) of the range of a flying Cobot and a rolling Shapeshifter, showing that it can reach between $100$ and $200$km. We observe that accounting for thermal and communication power requirements may partially reduce the range. Our analysis has been validated via a high-fidelity simulator, which takes into account the robot's dynamics, aerodynamics and control algorithms, based on a platform similar to our hardware prototype. Last, we have studied the optimal number of Cobots that should constitute a rolling Shapeshifter in order to maximize the range of the platform, concluding that the optimal number may be between 3  and 6 agents.

\chapter{Subsystems}
\label{sec:Subsystems}
In this section, we provide a first order estimate of the critical subsystems of each Cobot and for the Home-base.
This section is partially based on our related work \cite{shapeshifterUBuffalo}.

\section{Cobot}

\subsection{Power supply}
As presented in Section \ref{sec:RoboticPlatform}, actuating each Cobot may require $1-10$W depending on the mobility mode (flying or rolling). For simplicity, in Phase I, we do not consider the power required to swim in a liquid. Additionally, in Phase I, we estimate the power consumption to be $\approx 5$W for the onboard computers and cameras, $4$W for communication and $5$W for scientific instruments (sampling tool). From our thermal analysis, highlighted in the following section, we assume a total power required to maintain the inside of the Cobot at the minimum operating temperature of about $5$W. 
As described in Table \ref{table:batteryTechnology}, lithium-ion battery cells are the most suitable batteries for the mission considering their relatively high specific energy density and their relatively low operating temperature. Thanks to their wide market availability and relatively low-cost, lithium-ion battery cells are the ideal candidate to maintain the design of the Cobot as simple and as economical as possible.
Additionally, although Titan’s atmosphere at the Cobots’ operating altitude has an average temperature of $-179$C, the lithium-ion cells can be heated by using electrically powered heating coils. They are proven to work well in similar situations, as the Huygens probe used similar technology based on lithium-ion cells. The batteries of the Cobot can be recharged via the Home-base, which would host a docking mechanism adequate for power and heat transmission. 

\begin{table}[h]
\begin{center}
 \begin{tabular}{| r c c |} 
 \hline
 \textit{\thead{Battery \\ technology}} &  \textit{\thead{Minimum operating \\\ temperature }[C]} & \textit{\thead{Specific energy \\ density }[Wh/Kg]} \TBstrut\\
 \hline \hline
 Nickel-Cadmium & $-20$ & $25$-$30$  \TBstrut\\
 \makecell{\hfill Nickel-Hydrogen\\(Individual Pressure Vessel Design)} & $-10$ & $35$-$43$ \TBstrut \\
 \makecell{\hfill Nickel-Hydrogen\\(Common Pressure Vessel Design)} & $-10$ & $40$-$56$ \TBstrut \\
 \makecell{\hfill Nickel-Hydrogen\\(Single Pressure Vessel Design)}& $-10$ & $43$-$57$ \TBstrut \\
Lithium-Ion & $-40$ & $70$-$110$ \TBstrut \\
Sodium Sulfur & $-10$ & $140$-$210$ \TBstrut \\
 \hline
\end{tabular}
\end{center}
\caption{Battery options \cite{larson1992space} taken into account for the power supply system of each Cobot. }
\label{table:batteryTechnology}
\end{table}

Despite the currently low \ac{TRL}, promising research at JPL has also shown the feasibility of realizing batteries capable of operating at a temperature close to Titan's ambient temperature \cite{west2010sulfuryl}. Advances in the development of this and other ultra-low temperature technologies (e.g. electronic packaging \cite{del2018electronic}), may further simplify the Cobot's design.

\subsection{Thermal control}
The average temperature on the surface of Titan is about $-179$C, which is significantly lower than the minimum operating temperature of the batteries and other avionics that will be part of the Cobots. Therefore, a thermal design that retains and keeps the majority of heat produced by the Cobot subsystems inside the robot is crucial for the batteries to be able to operate. We assume the target internal temperature of the Cobot to be $0$C, in order to take into account the operating temperature of avionics and leave some safety margin for the minimum operating temperature of the Li-ion battery. In Phase I feasibility studies, we rely on passive methods to initially heat the Cobots before their initial disembarkment from their storage holds in the lander, and active methods, such as resistive heating elements, to maintain an internal temperature of $0$C. To minimize the total heat losses, a crucial role is additionally played by thermal insulation.
\subsubsection{Thermal insulation options}
\begin{table}[h]
\begin{center}
 \begin{tabular}{| r c c c |} 
 \hline
 \textit{\thead{Material}} &  \textit{Density $\frac{g}{cm^3}$} & \textit{Thermal conductivity  $\frac{W}{mK}$} & \textit{Density $\times$ Conductivity} \TBstrut \\
 \hline \hline
 Kapton Foil & $1.42$ & $.46$ & $.653$ \TBstrut\\
 Aerogel & $0.0019$ & $0.004$ & $.0000076$ \TBstrut \\
 Basotec Foam & $0.009$ & $0.0035$ & $.000315$ \TBstrut \\ 
 \hline
\end{tabular}
\end{center}
\caption{Thermal insulation options: Comparative Figures of Merit.}
\label{table:ThermalInsulationOptions}
\end{table}
Table \ref{table:ThermalInsulationOptions} lists various insulators that can be used on the Cobots to maintain the target internal operating temperature. Aerogel appears to be the optimal option, as it guarantees low conductivity and low mass, given a fixed volume available. Minimizing mass and volume of the system is essential to minimize thrust required for lift (due to mass) and body drag (due to volume).
\subsubsection{Thermal insulation analysis}
\begin{figure}
    \centering
    \includegraphics[width = 0.75\textwidth]{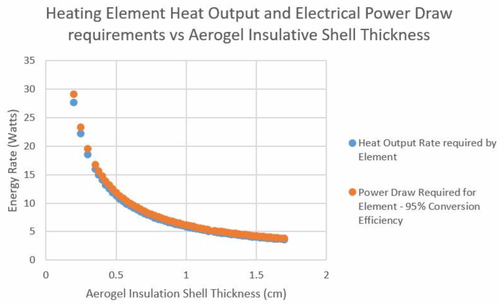}
    \caption{Active power required to maintain the internal temperature of $0$C in a Cobot as a function of the thickness of the aerogel insulating material. To simplify the analysis, we approximated the payload compartment of a Cobot as a spherical shell, with fixed internal radius of approximately $0.1$m.}
    \label{fig:insulator_vs_active_power_required}
\end{figure}

In this subsection, we provide a first order analysis of the mass and volume of aerogel required to sufficiently thermally insulate each Cobot. We also compute the power required by the internal resistance to compensate for the passive heat losses, maintaining the interior of the Cobot to the target temperature of $0$C.
\paragraph{Assumptions}
We approximate the cavity containing the electronics and the motors of the Cobot as a hollow sphere. Our goal is to compute the minimum thickness of aerogel needed to maintain the internal temperature of the cavity to around $0$C, with the conditions of the outer shell at constant freestream temperature of $-179$C. From Fourier's law, the 1-D radial conduction rate through a homogeneous substance is given by:
\begin{equation}
    Q_r = kA\frac{dT}{dr}
\end{equation}
where $Q_r$ [W] is the rate of heat transfer, $k$ is the conductive coefficient of the material, $A$ is the cross sectional area involved in the heat exchange. The expression $\frac{dT}{dr}$ defines the temperature ($T$) change with respect to radial position ($r$) in the material. 
For a thin-walled sphere we define $Q_r$ as:
\begin{equation}
    Q_r = \frac{4 \pi k r_1 r_2 (T_1-T_2)}{r_1-r_2}
    \label{eq:thermal_conductivity_sphere}
\end{equation}
where $r_1, T_1$ represent the internal radius and temperature, while $r_2, T_2$ the external radius and temperature, respectively. The scalar $k$ is the conductive coefficient of the aerogel, and $Q_r$ corresponds to the power that has to be produced by the active heating units inside the Cobot, assuming perfect thermal conductivity of the material inside the cavity. For simplicity, we neglect the effects of the atmospheric pressure on the aerogel (the effectiveness of the insulation decreases as the pressure increases) and we assume a thermal conductivity of $k = 0.004$W/mK.
\paragraph{Results}
From \ref{eq:thermal_conductivity_sphere} we can derive the necessary power $Q_r$ to be produced by the heating elements as a function of the thickness of the aerogel $r_2-r_1$, assuming an internal radius $r_1 \approx 0.1$m. The result is shown in Figure \ref{fig:insulator_vs_active_power_required}, where we have additionally considered the energy conversion losses of the active heating unit assuming an efficiency of $95\%$. From our analysis, we can allocate $\approx5$W to the active heating element for the electronics cavity in the Cobot, given a thickness of $\approx 0.01$m. This choice of the insulation option would require only $\approx 0.01$kg of mass of aerogel. 

\subsection{Tool for sampling}
\begin{SCfigure}
  \centering
  \caption{Tool for sampling mounted on the bottom of each Cobot.}
  \includegraphics[width=0.6\linewidth]%
    {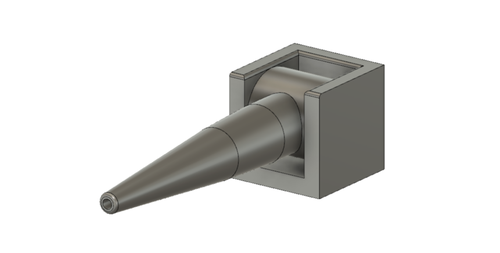}%
    \label{fig:sampler}
\end{SCfigure}

Each Cobot is equipped with a sampler, whose purpose is to collect small particles, fluids or gases for compositional analysis in the Home-base.
The sampler consists of a tapered tube attached to a motor that is mounted on the bottom of the Cobot. Inside the base of the tube there will be a vacuum to draw particles into the apparatus. The tip of the sampler will contain a seal that remains closed when not in use, and opens when the vacuum is activated. Furthermore, the end of the sampler is tapered to allow for stress distribution. Structurally, the apparatus will be able to withstand a direct impact with the surface in case of errors in the obstacle avoidance system. The sampler is represented in Figure \ref{fig:sampler}.
\par Future work will focus on detailing the design of the tool and on the servomechanism necessary to accurately control its position.

\section{On-board computation}
\begin{figure}
    \centering
    \includegraphics[width=0.6\linewidth]{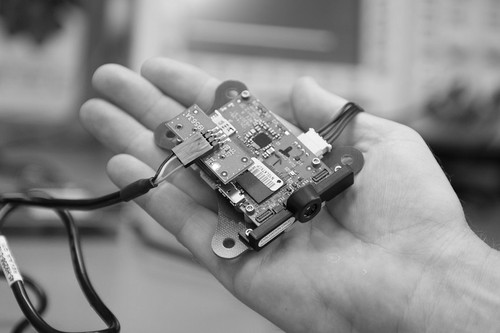}
    \caption{Qualcomm Snapdragon board, which includes two RGB cameras, a $2.1$GHz quad-core CPU and an \ac{IMU} in a compact form factor.}
    \label{fig:snapdragon}
\end{figure}

The reduced size and mass of the design of the Cobots poses a challenge in achieving the powerful on-board computation necessary to execute the autonomy algorithms (detailed in Section \ref{sec:Autonomy}). Adopting conventional high-performance computation platforms is unfeasible due to their power requirements, weight and size. A platform that can overcome these constraints, while providing sufficient computation power, is the integrated computer Qualcomm Snapdragon \cite{Snapdrag52:online}, shown in Figure \ref{fig:snapdragon}. It features a $2.1$GHz quad-core CPU, camera and \ac{IMU} in the form factor of a credit-card ($75$mm x $26$mm for the PCB), and wireless connectivity adequate for mesh networking.
\par
The Qualcomm Snapdragon has also been adopted for other space flight missions, such as the JPL's Mars Helicopter Scout \cite{balaram2018mars}. Qualcomm\footnote{\url{https://www.qualcomm.com} and JPL are currently collaborating on producing a radiation-hardened version of the product. Our team has demonstrated autonomy algorithms with this platform (\cite{qsf5}, \cite{qsf6}, \cite{qsf7}, \cite{qsf8}, \cite{qsf9}, \cite{qsf10}).}

\section{Home-Base}
\subsection{Power supply}
The lander will host the majority of the science instruments for the mission, as well as communication equipment to relay data back to Earth and the infrastructure to recharge the batteries of the Cobots. A high-output, high-reliability power source is therefore important for the platform. Due to the scarce solar irradiation, the lander would not be able to rely on solar energy, but rather on-board \ac{RTG}. 
\paragraph{Power requirement}
The exact power requirement for the lander depends mostly on the scientific and communication payload. An example of the payload considered in Phase I (and its respective power requirement) to dimension the electrical power subsystem for the lander is shown in Table \ref{tab:lander_science_payload}. A first-order estimate of the power necessary for the communication and scientific payload of the lander is of $\approx 250$W. An additional power demand should be factored-in in order to take into account the recharging of the batteries of the Cobots. Such value changes according to how often the Cobots are employed in a task, and for simplicity we will assume that the excessive power produced by the lander will be used to recharge the Cobots. 
\begin{table}[]
    \centering
    \begin{tabular}{|r|c|}
    \hline
         \textit{Subsystem} & \textit{Estimated Power} [W]  \\
    \hline
    \hline
        Computer    &   10  \\
        PIXL \cite{Planetar20:online}           &   25  \\
        Chromatograph \cite{Chromagr91:online}  &   170 \\
        Communication & 5  \\
        \hline
        \multicolumn{1}{r}{\textbf{Total:}} & \multicolumn{1}{c}{$\approx 200$W} \\
        \multicolumn{1}{r}{\textbf{Total + 25\%:}} & \multicolumn{1}{c}{$\approx 250$W} \\
    \end{tabular}
    \caption{Power requirement for the lander. Excess power goes to Cobot charging.}
    \label{tab:lander_science_payload}
\end{table}

\paragraph{Power source}
In Phase I, we do not consider conventional energy harvesting methods for mission power generation while on Titan due to the scarce solar irradiation on the moon. Our power system would be instead based on a radioisotope generator such as a \ac{MMRTG}. 
\par A promising radioisotopes-based power source that Shapeshifter could use is the Modular Stirling Radioisotope Generator (MSRG) \cite{Microsof36:online}. Significant work remains to raise this technology to be flight ready, but it has the potential to be smaller, lighter, more robust, and generate more power per kilogram of system mass than a conventional RTG  \cite{Microsof36:online}. Depending on the size of the engine and the isotopic samples utilized, it can provide about $50$W to $450$W to the spacecraft \cite{Microsof36:online}. Such technology, thanks to the cycle employed and the temperature differential with Titan’s atmosphere, would produce power with increased efficiency than conventional RTGs. Unfortunately, as the pistons are by design not inline, some of the more reliable configurations of the MSRG would require the lander to dampen the vibrations produced by the power system, and should be addressed in future works. According to our analysis and \cite{Instrume93:online}, a configuration with 6 modules would be optimal for our power requirements. Such configuration would produce a power output of $369$W in its beginning of life, and would only drop to about $303$W $17$ years later. This configuration, given its mass of $103$kg, would output a specific power of $3.6$ W/kg. The MSRG should be assisted by an auxiliary battery that provides the necessary peak-power for a quick recharge of the Cobots. Dimensioning and technology of the battery are left as future work.

\subsection{Thermal control}
\paragraph{Thermal insulation} We envision to inspire the thermal insulation design of the Home-base from the Huygens’ probe \cite{clausen2003huygens}. A detailed analysis is left as future work. %

\paragraph{Heat generation and control}
The Home-Base will be equipped with heat sinks to directly transfer and evenly distribute the heat generated by the \ac{MSRG} power generator. Additionally, the internal temperature may be regulated via Radioisotope Heater Units (RHUs)  strategically positioned near critical electronics equipment. Excessive heat could be dissipated via venetian-blind louvers or transported in different areas of the lander via heat pipes. A detailed analysis of the thermal control of the lander will be addressed in future works. 

\subsection{Charging bays for the Cobots}
\begin{figure}
    \centering
    \includegraphics[width=0.8\linewidth]{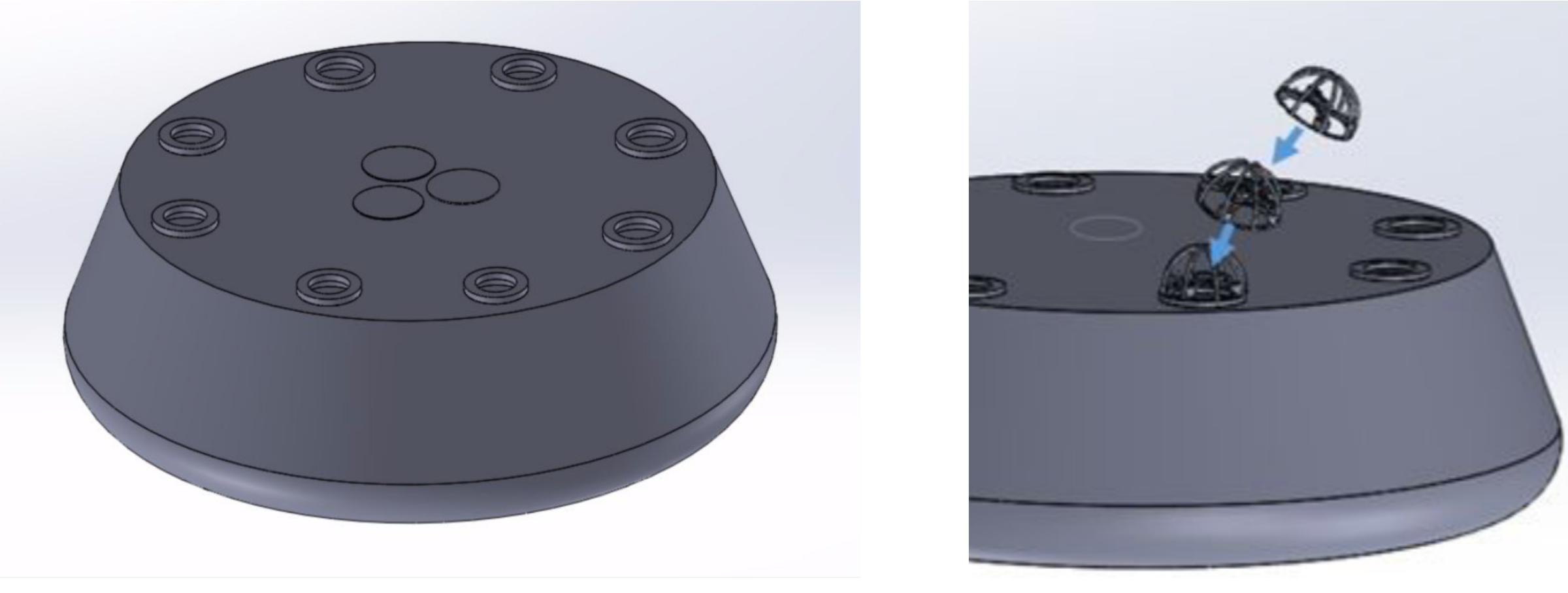}
    \caption{\textit{(Left:)} Charging bays installed in the Home-base. \textit{(Right:)} Example of docking manoeuvre of a hemispherical Cobot in the charging bay of the Home-base.}
    \label{fig:charging_bays}
\end{figure}

The Home-base will contain a charging station where the Cobots will be able to recharge their onboard batteries. The Cobots will be able to dock to the charging station via the same magnets used to dock with other Cobots. Beyond power, the charging station will be able to transfer to the Cobots the excessive heat produced by the Home-base. A conceptual representation of the charging bays for the Cobots and the docking maneuver is represented in Figure \ref{fig:charging_bays}. 

\section{Communications}
\subsection{Hardware}
Communication with Earth will rely on a combination of directional high-gain and low-gain antennas similar to what was used for the Cassini Mission. Such antennas will be installed on the Home-base, which offers enough power to establish a reliable link with Earth. The Home-Base will additionally host an antenna to communicate with each or some of the Cobots. Each Cobot will host a compact antenna (e.g. dipole) to communicate with the Home-base and to communicate with other Cobots.

\subsection{Software}
The autonomy framework of the Shapeshifter (described in Section \ref{sec:Autonomy}) relies on the ability of the Cobots to communicate with each other and with the Home-base. This is used, for example, for increased safety in the exploration of underground environments, to off-board computationally intensive tasks to the Home-base, or to more efficiently map large areas.
\subsubsection{Mesh networking}
\begin{figure}
    \centering
    \includegraphics[width=0.8\textwidth]{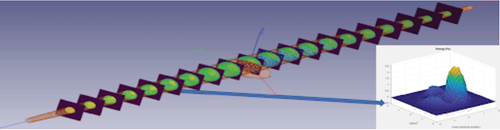}
    \caption{Wave-guide effect in underground cave visualized with communication modelling. The bouncing of the electromagnetic waves on the walls of the cave enables the communication between two nodes that do not have line of sight.}
    \label{fig:autonomy_comm_node_simulation}
\end{figure}

To increase the robustness of the communication channel and avoid creating a single point of failure typical of centralized communication approaches, we employ a decentralized infrastructure based on a mesh-networking.
In a mesh network, each node (e.g. a Cobot or the Home-base) is capable of communicating with any other neighboring agent and contributes in sending and receiving messages to any other node in the network. 
Recent work \cite{walsh2018communications} has also highlighted the possibility of modeling communication in constrained and cluttered environments, allowing to predict the optimal placement of the communication node of the network in caves and canyons. An example of the propagation of the communication carrier in a cave is represented in Figure \ref{fig:autonomy_comm_node_simulation}.

\paragraph{Preliminary mesh-networking experimental results}
In our feasibility experiments, we have created a mesh-network between a ground station and a mobile robot, deploying $4$ communication nodes, spaced about $40$m each, that guarantee a robust communication channel between the two communication endpoints, achieving $\approx 700$Kbps.

\subsubsection{Disruption Tolerant Networking}
The mesh-network is additionally equipped with the \ac{DTN} \cite{burleigh2003delay} technology, which stores and forwards messages in case of communication lost between one node and the network. \ac{DTN} addresses the inevitable intermittent nature of connectivity in hard-to-reach environments such as canyons, mountains, and cave networks.

\section{Radiation protection considerations}
\begin{figure}
    \centering
    \includegraphics[width=0.65\linewidth]{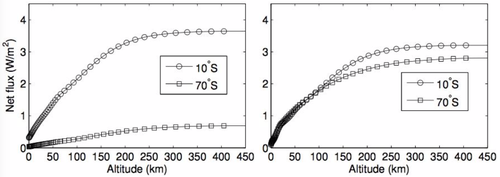}
    \caption{Solar Flux vs Titan Altitude. Graphs showing the new solar flux in short-wave (left) and long-wave (right) diurnal-means for different altitudes during Titan equinox \cite{lora2014radiation}.}
    \label{fig:solarFluxAtTitan}
\end{figure}

Thanks to Titan's thick atmosphere and to its large distance from the Sun, Titan's surface is well protected from cosmic rays. As shown in Figure \ref{fig:solarFluxAtTitan}, the majority of the cosmic rays that reach Titan don’t make it past the upper atmosphere, and the remaining $90\%$ of them is blocked by the haze.
For approximately 95\% of the time, extra protection from cosmic rays is also provided by Saturn’s magnetosphere \cite{Overview20:online}. Since the Cobots’ primary mission will take place on or near the surface of Titan, we don't consider radiation protection for the Cobots and the Home-base as a  critical aspect that may undermine the feasibility of the mission.

\section{Summary}
In this section, we have analyzed critical subsystems (power, thermal insulation and control, and communication) for each Cobot and for the Home-Base. Our preliminary analysis indicates the most of the subsystems could be implemented with existing or high-readiness technology. Future work should focus on detailing the thermal and communication strategies for the Cobot, while we have shown that solar radiation will not be a major issue for the survival of the Shapeshifter on Titan. The adequateness of the employed subsystems for under-liquid operations will also be addressed in future works.

\chapter{Autonomy}
\label{sec:Autonomy}
 Some of the more interesting places for current life-supporting conditions are in difficult-to-reach places such as caves, volcanoes and lake boundaries: areas for which Shapeshifter is specifically suited. Due to the communication delay between Earth and Titan, which is larger than one hour, any teleoperation or remote decision making for the execution of the Shapeshifter mission is unfeasible. Full-autonomy, with no provision of human control, is therefore one of the most critical requirements of the Shapeshifter mission. In this section, we investigate the feasibility of the main components necessary to achieve full autonomy; such feasibility-analysis builds upon several ongoing works in our group at NASA's Jet Propulsion Lab or previous work from our team members.

\section{Mapping and localization}
The Shapeshifter, in any of its configurations (Flight Array, Rollocopter, Torpedo), relies on accurate localization (i.e. positioning) capabilities in order to navigate in potentially labyrinthic environments like canyons, faults, caves and the cryolava tubes that may be found on Titan. A dense, 3D map of the explored environments will be sent back to Earth via the Home-base and will be used to identify new areas to explore, as well as locate large geological features present on Titan. Mapping and localization services will benefit from increased robustness and efficiency offered by multiple Cobots that collaboratively build a map of the environment. 
\subsection{LAMP}
\begin{figure}
    \centering
    \includegraphics[width=0.70\textwidth]{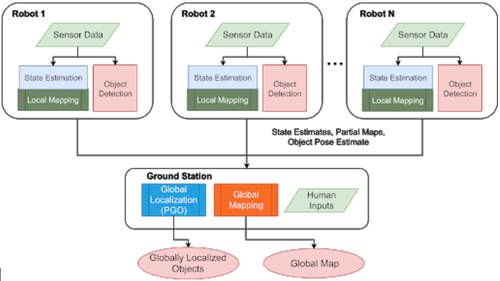}
    \caption{Conceptual diagram of the framework for localization and mapping based on Distributed Pose Graph Optimization.}
    \label{fig:autonomy_multirobot_mapping}
\end{figure}

\ac{LAMP} is our proposed framework to meet the multi-agent localization and mapping requirements of the mission. \ac{LAMP} is based on Distributed Pose Graph Optimization (DPGO). It is able to provide accurate global localization and large scale topological and geometrical 3D maps of above-surface and subterranean networks through the inter-agent sharing of sparse graphs and multiple loop-closing modalities. To study the feasibility of the proposed mapping method, a conceptual diagram of the framework is represented in Figure \ref{fig:autonomy_multirobot_mapping}, and the map produced by \ac{LAMP} (in a $\approx 200$m-long subsurface void, which may provide a good test scenario to simulate a cave on Titan) is represented in %
Figure \ref{fig:autonomy_blam_detail}.

\subsection{Efficient mapping for robust planning and long-term operations}
The pointcloud-based 3D map generated by \ac{LAMP} provides useful data on the morphology and topology of the explored area, but it does not explicitly embed information on the quality of the map, which may vary due to sensor noise and measurement uncertainties; such information is necessary for robust planning and obstacle avoidance. To overcome these limitations, each Cobot will additionally generate a confidence-rich map, following our previous work \cite{agha2017confidence} and \cite{agha2019confidence}. Confidence-rich maps will then be used for local/global planning and collision avoidance.
\par
For agile or emergency operations, such as rapid escape maneuvers near obstacles, Cobots will employ a holistic framework \cite{morrell2016integration}, \cite{morrell2018enhancing}, which leverages combined mapping, obstacle avoidance, and 3D trajectory generation to guarantee performance levels not achievable by conventional, de-coupled approaches.
\par 
The long-term/wide-range operations envisioned for the Shapeshifter, coupled with the constrained on-board computational resources, poses challenges on data summarization and data priorities in order to generate an adequate representation of the environment. Following the map-reduction approach proposed in \cite{mu2015two}, the Cobots will generate a resource-efficient map by taking into account task-specific requirements, allowing them to focus their attention on certain features or part of the environment that best allow them to accomplish their goals (e.g. navigation, sample collecting). 

\subsection{Autonomous identification of features of interest and opportunistic science}
We are currently exploring machine learning based semantic understanding integrated within the Distributed Pose Graph Optimization framework to provide the global location of specific topological or geological features in the constructed map, such as canyons, rivers, lakes, or junctions in cave networks. {Learning-based methods are also used to provide estimates of terrain traversability in the extreme environments. In Figure \ref{fig:autonomy_junction_detector}, we show the feasibility of detected junction in a simulated cave network as a representative example of the feasibility of this approach in a simple setting.
Learning-based approaches will additionally allow for automatic detection of novel science opportunities, such as geological features \cite{kerner2019novelty}, which will be used to autonomously define tasks for the mission.

\begin{figure}
    \centering
    \includegraphics[width=0.9\textwidth]{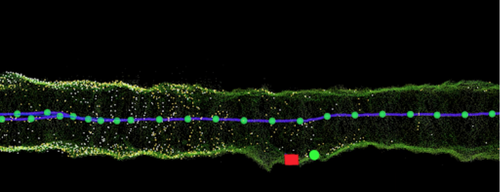}
    \caption{Detail of the map generated by LAMP when moving in straight subsurface voids. }
    \label{fig:autonomy_blam_detail}
\end{figure}

\begin{figure}
    \centering
    \includegraphics[width = 0.5\textwidth]{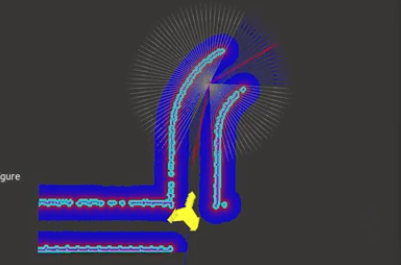}
    \caption{Junction detected (in yellow) in a simulated cave network, used for mission planning and dispatching Cobots to various part of the terrain}
    \label{fig:autonomy_junction_detector}
\end{figure}

\section{State estimation}
Shapeshifter's navigation requires a reliable state estimate through different scenarios, such as under-liquid \cite{garcia2017exploring}, subterranean or low-lit environments. The presence of obscurants, like dust and methane fog, make state estimation even more challenging. We propose to increase the robustness of the estimation framework in two ways, by making use of different sensors with different failure modalities, and by actively modeling failure modes, defining recovering strategies for each scenario. 

\subsection{Multimodal (Visual, Thermal, Laser, Inertial) state estimation}
Multimodal state estimation combines complementary sensing modalities to guarantee sufficient performance in a range of challenging sensing scenarios. We are currently assessing the feasibility and the performance of two main pose-estimation configurations for a single Cobot (and, as a consequence, for the Shapeshifter in its different configurations):
\begin{itemize}
    \item Visual-Inertial odometry, for navigation in well-lit environments. It relies on the \ac{IMU} and on the stereo camera that is mounted on each Cobot. Stereo images are used to detect and track 2D visual features (i.e. corners, blobs, ORB-features \cite{rublee2011orb}) that are used to estimate the motion of the robot.
    \item Laser-Inertial odometry, for navigation in underground, poorly illuminated environments, or for navigation in obscurants such as methane fog and dust. Uses an \ac{IMU} and 3D features detected in the point-cloud produced by a LiDAR to provide an estimate of the pose and velocity of each Cobot. 
\end{itemize}
\begin{figure}
    \centering
    \includegraphics[width=\linewidth]{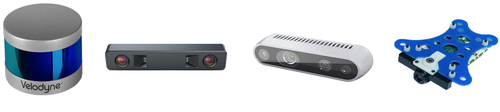}
    \caption{Sensors considered to evaluate different sensing modalities to check feasibility of using each sensor. This analysis will allow us to chose a specific payload for each Cobot. \textit{From left to right:} Velodyne VLP-16 LiDAR (source: \href{https://www.velodynelidar.com/puck/}{https://www.velodynelidar.com/puck/}), Minteye Stereo Camera (source: \href{https://www.mynteye.com/products/mynt-eye-stereo-camera}{https://www.mynteye.com/products/mynt-eye-stereo-camera}), Intel-Realsense D435 RGBD camera (source \href{https://www.intelrealsense.com/depth-camera-d435/}{https://www.intelrealsense.com/depth-camera-d435/}), Qualcomm Snapdragon Flight monocular camera source (\href{https://developer.qualcomm.com/hardware/qualcomm-flight-pro}{https://developer.qualcomm.com/hardware/qualcomm-flight-pro}).}
    \label{fig:autonomy_hand_held}
\end{figure}

We explicitly take into account the possibility of failure of each state estimation approach by introducing a ``switching" logic%
which runs multiple parallel estimators based on various modalities, and continuously switches between the most reliable state estimator. The optimal state estimate is chosen according to a set of confidence tests which are based on criteria such as the covariance of the state estimate and the frequency of its output. 
Figure \ref{fig:autonomy_hand_held} shows various sensors used to compare different state estimation modalities.

\section{Motion planning and multi-agent coordination}

In this part, we present motion planning approaches and the feasibility of the complex tasks that the Shapeshifter has to accomplish.
\subsection{Motion planning under uncertainty}
A fundamental feasibility concern for the Shapeshifter is the ability to efficiently make complex decisions under uncertainty. In order to operate at its full capabilities, for example, the Shapeshifter will face the constant need to decide in which configuration to morph, based on multiple factors such as the environment (e.g. terrain), its battery level or the goal of the mission. Inevitable imperfections in the sensing component of the Cobots (e.g. noise on the raw sensor input, calibration errors, unexpected anomalies occurring on Titan) or in the motion of the robot (e.g. due to manufacturing imperfections on the actuators or on the propellers) can affect the ability to take the correct decision, and can quickly deteriorate the robustness of the platform if not explicitly taken into account in the planning phase.
\par 
In its most general form, a planning problem where uncertainty is explicitly taken into account can be formulated as a \ac{POMDP}. Uncertainty is encoded in a \ac{MDP} (which describes the system's dynamics) by maintaining a probability distribution (called \textit{belief}) over the set of possible states, where the probability depends on the observations (e.g. sensor measurements) and observation model (the type of sensor). The POMDP approach is able to guarantee an optimal solution (called \textit{policy}) of the planning problem, capturing multiple sources of uncertainty, and is therefore employed for the motion planning task of the Shapeshifter. Its complexity, however, makes finding the optimal policy computationally challenging or often intractable on small platforms in real-time.
\par
To handle this challenge and address the feasibility and computational challenges of planning with \acp{POMDP}, we rely on \ac{FIRM} \cite{agha2014firm}. FIRM provides a tractable and risk-aware methodology for solving the problem of motion planning under uncertainty by reducing the problem to a tractable one on a representative graph capturing the traversable space for Cobots. \ac{SLAP} \cite{agha2015simultaneous} and \cite{kim2019bi} extends the work on \ac{FIRM} by proposing a dynamic replanning scheme in belief space that enables online replanning, as the new elements of the environment are discovered. Such an approach leads to intelligent robot behaviors that provably takes the solution closer to the optimal solution and guarantees that success probability only increases.

\subsection{Trajectory generation}
Once high-level waypoints are decided by the POMDP solver, a trajectory generator is needed to take the robot from one reference position to the next.
Different methods can be employed to generate a dynamically feasible trajectory for the Shapeshifter. The approach proposed in \cite{otsu2017look}, \cite{janson2015fast}, \cite{ichter2017real} will allow to generate a dynamically-feasible motion plan subject to a constraint on collision probability, and will allow to best tradeoff the competing objectives of performance and safety, for example depending on the level of risk that the Shapeshifter mission is able to accept at its different stages.  The method described in \cite{tagliabue2019energy} will instead guarantee that the Shapeshifter will always operate at the most energy-efficient velocity, regardless of unexpected factors such as actuator degradation or payloads, by finding online the velocity which maximizes the range or the endurance of the robot.

\subsection{Multi-agent coordination}
Cooperation between Cobots and Home-base will be necessary in order to accomplish the complex mission tasks and shapeshifting behaviors of the platform. Recent work has shown the feasibility for the Cobots to cooperate on many different levels and while morphed in different configurations. Example include cooperation between:
\begin{itemize}
    \item Flying Cobots and Rollocopter: similar to what proposed in \cite{nilsson2018toward}, a group of Cobots morphed in a Rollocopter will explore an unknown environment assisted by a swarm of flying Cobots so as to maximize knowledge about a science mission expressed in linear temporal logic. This has shown to lead to intelligent decisions even in case of uncertainty \cite{nilsson2018toward}. 
    \item Flying Cobots and Home-base: The approach \cite{collaborativeTransportation}, \cite{tagliabue2017collaborative} will instead allow for robust collaborative transportation of the Home-base using multiple Cobots. The approach guarantees high levels of robustness as it has the advantage of not requiring any explicit communication channel (which may be subject to delays or failures) between the Cobots.
    \item Flying Cobots will execute path-planning tasks \cite{pavone2007decentralized} in a fully decentralized fashion, for increased robustness to system failure (loss of one or more Cobot) or communication failures.
\end{itemize}

\section{Task and mission planning for distributed systems}
The multi-agent multi-modal nature of the Shapeshifter creates complexity and challenges in terms of mission planning. In order to maintain consistently high levels of productivity for the missions, Shapeshifter needs to rely on an automated, multi-agent decision framework which makes use of only high-level guidance with very low-frequency from a human operator on Earth to select its own situational activities. Further, it needs to respond to unexpected conditions, without dependence on ground intervention. Recent work (e.g. \cite{gaines2018self}) has studied the feasibility of single-agent solutions for multi-day task planning for spacecraft/rovers, and our aim is to present approaches that make the automated mission planning for the Shapeshifter feasible.
\par

A framework that shows the feasibility of archiving the autonomous task planning goal while guaranteeing robustness and awareness to uncertainty is based on \ac{Dec-POMDP}. As previously discussed, POMDP-based approaches present the advantage of explicitly taking different sources of uncertainty into account to achieve increased resilience. Dec-POMDPs extends the single agent \ac{POMDP} formulation to a multi-robot setting.
Despite its power, this framework presents computational challenges that make it unfeasible on resource-constrained systems.
\par
 Recent work (e.g., \cite{omidshafiei2015decentralized}) has studied the feasibility of overcoming such computational challenges by introducing a novel formulation, the Decentralized Partially Observable Semi-Markov Decision Process (Dec-POSMDP), which additionally allows for asynchronous decision-making by the robots, crucial in multi-robot domains.
\par 
Different efforts have also extended the situational awareness of a robot to the point that it can better understand its environment and its own state. 
The work in \cite{ure2015online}, for example, enables the online estimation of the unknown state transition models associated with the uncertainty that stems from mission dynamics. This will allow the Shapeshifter to better adapt to Titan's challenging environment, for example by taking decisions according to its understanding of wind and currents and how they are affected by nearby geological features. \cite{agha2014health}, instead, shows the feasibility of taking into account the system’s health and capability degradation, an essential factor to predict and avoid failures in persistent missions.

\chapter{Mission architecture to Titan}
 \label{sec:MissionOperationsTitan}
In this section, we present a preliminary analysis for a Titan mission scenario. We will especially focus on the exploration of the potential cryovolcano Sotra Patera, and its unique geological features. 

\section{Science objectives at Titan}
Liquid water and organics are essential for life but, despite being commonly found throughout the Solar System, locations where they are known to coexist are rare. Titan’s cryovolcanic regions are high priority locations to search for contact between liquid water and complex organic material because those environments would be habitable for the period of liquid water persistence. The Sotra Patera region, illustrated in Figure \ref{fig:SotraPateraMissionArchitecture}, is the strongest candidate cryovolcanic feature on Titan \cite{lopes2013cryovolcanism}. Shapeshifters will explore the Sotra Patera region to confirm its cryovolcanic origin and determine the extent that liquid lavas have interacted with organic surface materials. Shapeshifter's underwater capabilities will also allow it to explore under-liquid environments such as Ligeia Mare, represented in Figure \ref{fig:LigeiaMare}.
\begin{figure}[h]
\centering
\includegraphics[width=0.6\columnwidth]{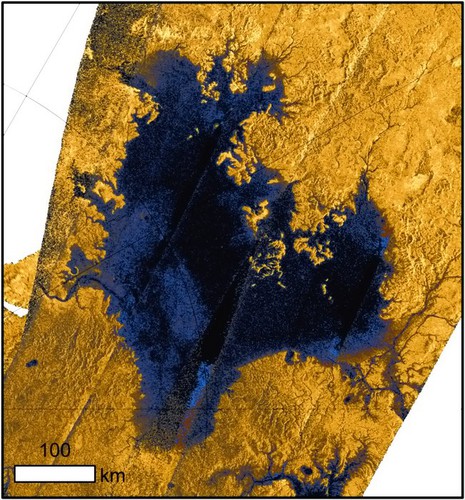}
      \caption{Ligeia Mare, one of the environments on Titan accessible to the Shapeshifter.}
\label{fig:LigeiaMare}
\end{figure}

\section{Science payload and platform's design}
In this part, we detail the science payload and the platform design tailored for a mission to Titan and the cryovolcano Sotra Patera. 
\subsection{Cobot}
Each individual Cobot will carry an optical stereo camera for the purposes of navigation and scientific imaging of Titan’s morphologies. The Cobots’ ability to image from the surface to high altitudes will allow significant flexibility in image resolution and coverage. Due to size and energy limitations, Cobot may carry only small and low-power instrumentation, such as the equipment typical considered for small spacecraft/rover hybrids (e.g. \cite{pavone2013spacecraft}), which include \acp{IMU}, thermocouples and a radiation monitor.  Each Cobot will be additionally equipped with a suction-based sample collecting unit, to collect samples of rocks or liquids to be analyzed by the Home-base. As summarized in Table \ref{tab:cobot_mass_and_power}, each Cobot will have a mass of approximately $1.1$kg, and will use an average power of $31$W. From our prototype, we additionally estimate that each Cobot will have a size of approximately $0.4 \times 0.4 \times 0.2$m, yielding to an expected volume of $0.03$m$^3$.
According to our analysis, the optimal number of Cobots used to create a spherical Shapeshifter (Rollocopter) will be 6, and thus we will carry at least 6 Cobots to Titan. More Cobots can be added easily added for redundancy according to the payload capabilities of the spacecraft carrying the Shapeshifter to Titan. 
If the current trend of increasing miniaturization for LiDARs will continue, Cobots will be additionally equipped with a mini-LiDAR (of weight below $300$g) for increased navigation resilience via multi-modal state estimation.%

\begin{table}
    \centering
    \begin{tabular}{|r|r|l|l|}
    \hline
    \textit{Category} & \textit{Instrument} & \textit{Mass} [g] & \textit{Power} [W] \\
    \hline 
    \hline
    Science Package & Sampling Tool & $50$ & $1$ (avg.) \\
    Science Package & Thermocouple & $50$ & $1$ \\
    Science Package & Radiation Monitor & $30$ & $0.1$ \\
    Autonomy & Autopilot w/ IMU & $30$ & $1$ \\
    Autonomy & Companion Computer & $70$ & $3$ \\
    Autonomy & Stereo Cameras & $20$ & $1$ \\
    Subsystems & Battery & $100$ & - \\
    Subsystems & Passive Thermal Insulation & $50$ & - \\
    Subsystems & Active Thermal Control & $10$ & $5$ \\
    Subsystems & Antenna and Transceiver & $40$ & $4$ \\
    Structural & Frame & $250$ & - \\
    Structural & Permanent-Electromagnets & $40$ (total) & $0.1$ (avg.) \\
    Structural & Actuators & $160$ (total) & $10$ (avg., conservative) \\
    \hline
    \hline 
    \multicolumn{2}{r}{\textbf{Total:}} & \multicolumn{1}{l}{$\approx 900$g} & \multicolumn{1}{l}{$\approx 25$W} \\
    \multicolumn{2}{r}{\textbf{Total + 25\%:}} & \multicolumn{1}{l}{$\approx 1100$g} & \multicolumn{1}{l}{$\approx 31$W} \\
    \end{tabular}
    \caption{Baseline design for a Cobot. The total mass of a Cobot would be $\approx 1$ kg and its average power requirement would be $\approx 30$W, where actuators and thermal control take most of the energy used by the platform. The enclosure would have, approximately, $0.2$m of radius. The power estimates are computed in Section \ref{sec:Subsystems} and are here repeated for clarity.}
    \label{tab:cobot_mass_and_power}
\end{table}

\subsection{Home-base}
The Home-base will host most of the scientific payload to analyze the samples that the Cobots retrieve; it will additionally act as a lander for the Cobots during the Entry, Descent and Landing (EDL) phase to Titan. Its design is inspired by Cassini’s Huygens lander.
Sample analysis capabilities are essential for satisfying the secondary objectives for finding life and studying organic chemical reactions that occur on the surface. Also, understanding the composition of the surface thoroughly will provide information on the early formation of celestial bodies, considering Titan is still in its early stages of development. A reasonable candidate to accomplish this task is PIXL from the Mars 2020 mission, as it is already designed completely and is capable of performing the necessary analysis on extraterrestrial planets. Additionally, a gas and liquid chromatograph will be included in the lander. Cobots will be able to sample fluids as well and return them to the lander for analysis, as PIXL is meant for soil. Table \ref{tab:homebase_mass_and_volume} collects the mass and volume that we predict for the Home-base.  
\par Using the Shapeshifter for sample collection and the Home-base for in-situ analysis is an optimal solution because the Shapeshifter can access any terrain while the in-situ instrumentation is not subject to the rolling and accelerations required for acquiring difficult samples. This combination allows for unconstrained sample acquisition and a more sensitive sample analysis. 

\begin{table}
    \centering
    \begin{tabular}{|r|r|l|l|l|}
    \hline
    \textit{Category} & \textit{Instrument} & \textit{Mass} [kg] & \textit{Volume} [cm$^3$] & \textit{Power} [W]\\
    \hline 
    \hline
    Science Package & PIXL & $6.9$ & $13400$ & $25$ \\
    Science Package & Chromatograph & $50$ & $144000$ & $170$\\
    Autonomy & Computer & $1$ & $10$ & $10$\\
    Autonomy & GNC sensors for EDL & $1$ & $50$ & $5$\\
    Subsystems & MSRG & $103$ & $87000$ & -\\
    Subsystems & Aux. battery & $11.4$ & $1150$ & - \\
    Subsystems & Antenna and Transceiver & $40$ & $4$ & $5$ \\
    Structural & Frame, docking and thermal & $\approx 100$ & - & -\\
    \hline
    \hline 
    \multicolumn{2}{r}{\textbf{Total:}} & \multicolumn{1}{l}{$\approx 300$kg} & \multicolumn{1}{l}{$\approx 245000$cm$^3$} & \multicolumn{1}{l}{$\approx 200$W} \\
    \multicolumn{2}{r}{\textbf{Total + 25\%:}} & \multicolumn{1}{l}{$\approx 375$kg} & \multicolumn{1}{l}{$\approx 300000$cm$^3$} & \multicolumn{1}{l}{$\approx 250$W}\\
    \end{tabular}
    \caption{Baseline design for the Home-base. The total mass of the Home-base will be $\approx 375$kg, its volume would be $\approx 300000$cm$^3$, and its power requirement will be on average of $250$W. Excessive power will be used to recharge the Cobots.}
    \label{tab:homebase_mass_and_volume}
\end{table}

\subsection{Mothership}
The Shapeshifter platform (Home-base and the Cobots) will be carried to Titan's orbit by a spacecraft, called the Mothership, which will resemble the spacecraft Cassini, whose CAD rendering is represented in Figure \ref{fig:cassini_cad}. The strong resemblance between the Huygens probe (which was about $1.3$m of diameter, and weighed roughly $318$kg) and the Home-base make a design inspired by the Cassini mission adequate for our mission.

\begin{figure}
    \centering
    \includegraphics[width=0.6\textwidth]{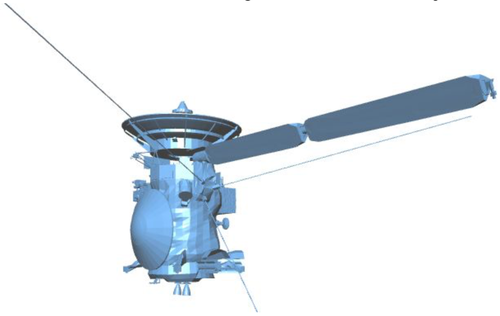}
    \caption{CAD rendering of Cassini. Shape and size of the Home-base are comparable to the Huygens probe carried by Cassini, and a design inspired by this spacecraft will be used to carry the Shapeshifter platform to Titan. Source: \url{https://nasa3d.arc.nasa.gov/detail/jpl-vtad-cassini}}
    \label{fig:cassini_cad}
\end{figure}

\subsection{Packaging of the Cobots}
During the travel to Titan and the EDL to the satellite, the Cobots may be stored in a compartment inside the Home-base, as represented in Figure \ref{fig:packaging_inside_homebase}. Further evaluation of packaging options (such as the storage rack represented in Figure \ref{fig:packaging_multi_rack}) is left as future work. 

\begin{figure}
\centering
\begin{minipage}{.47\textwidth}
  \centering
  \includegraphics[width=.9\linewidth]{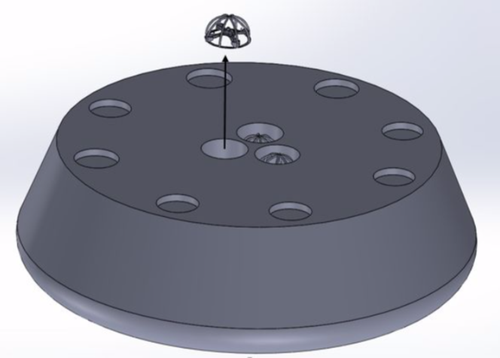}
  \captionof{figure}{Example of packaging of the Cobots inside the Home-base.}
  \label{fig:packaging_inside_homebase}
\end{minipage}%
\hfill
\begin{minipage}{.47\textwidth}
  \centering
  \includegraphics[width=.9\linewidth]{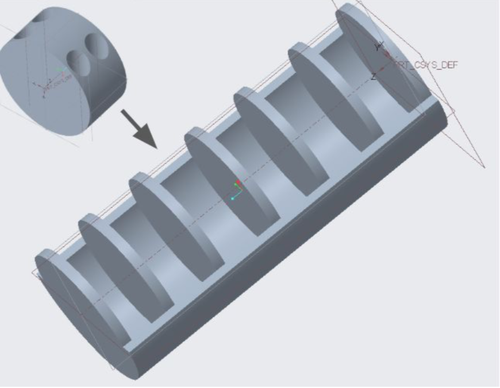}
  \captionof{figure}{Example of packaging of Cobots in an external rack.}
  \label{fig:packaging_multi_rack}
\end{minipage}
\end{figure}

\section{Case study to Titan's possible Cryovolcano Sotra Patera}
\subsection{Entry, descent and landing sequence}
Due to the similarities of the Home-base and the Huygens probe, the EDL phase is analogous. The Home-base lander will be assisted by an entry and descent module, which carries mission control components, a heat shield for deceleration, thermal protection, and a parachute. 

\subsection{Exploration of Sotra Patera and other unique features of Titan}

\begin{figure}
\centering
\includegraphics[width=\linewidth]{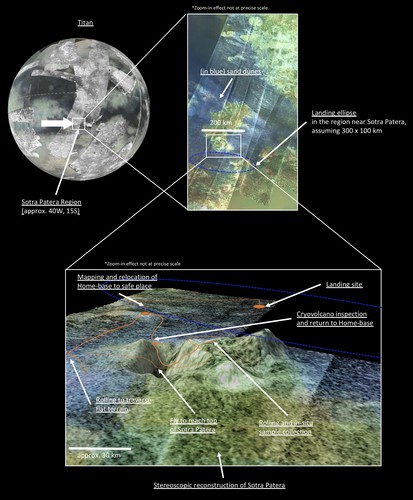}
      \caption{Example of mission scenario near Sotra Patera. The landing area is chosen in the proximity of the cryovolcano Sotra Patera, and we assume a landing ellipse of 100 km by 300 km. After landing, the Shapeshifter maps the surrounding environment and relocate the Home-base to a safe place. From there, the Shapeshifter moves, by rolling or flying, to the summit of the volcano, where in-situ samples are collected. After shapeshifting into a Rollocopter to inspect the caves near the cavity of the volcano, Shapeshifter returns to the Home-base to analyze the collected samples and plan the next mission.}
\label{fig:SotraPateraMissionArchitecture}
\end{figure}
Our preliminary landing location is near Sotra Patera, the most likely site of cryovolcanism on Titan, where our portable Home-base and a Shapeshifter, as a collection of $48$ or more Cobots, are deployed. Once deployed, Shapeshifter will begin to morph into the optimal configuration based on the observed properties of the terrain. It will start by building high-resolution terrain maps of the region near their base. Then, it will continue shape-shifting to traverse on the Titan’s surface, gather and relay science to the home-base. Examples of the science capabilities of the platform are:
\begin{itemize}
    \item \textbf{Low/high-resolution local mapping}: Shapeshifter builds maps of the region near their base; the trade-off accuracy vs time to map can be tuned by varying the flight altitude of the Cobots. 
    \item \textbf{Stratigraphy, fault survey and surface conductivity survey}: Shapeshifter explores cliffs and faults to analyze their potential sedimentary nature and measure the conductivity of the surface;
    \item \textbf{Deep excursion}: Shapeshifter explores to its maximum range, making the most efficient use of available energy by switching between the flight array and Rollocopter modes;
    \item \textbf{Cave exploration}: Shapeshifter explores detected caves and cryolava tubes in the Rollocopter mode  and comm-mesh-chain mode;
    \item \textbf{Mare diving for bathymetry and composition survey}: Shapeshifter morphs into a swimmer to dive under the surface of Titan’s mare, collecting samples and creating a 3D map of the surrounding environment;
    \item \textbf{Active/passive seismometry}: basic seismography studies can be performed using the onboard accelerometers used for GN\&C, while the Home-base can host a seismometer.
\end{itemize}
 After observing and analyzing a science site, Shapeshifter rebases, i.e., it morphs into a transporter and move the Home-base to a new mission site (see \ref{fig:SotraPateraMissionArchitecture} \textit{(right)}). An example of our mission is represented in Figure \ref{fig:SotraPateraMissionArchitecture}.

\subsection{Mission duration}
We envision a mission of the total duration of two Titan days, equivalent to approximately 31 Earth days.

\chapter{Conclusions}
\label{sec:ConclusionAndFutureWorks}
In this work, we have presented a novel, multi-agent, multi-modal, self-assembling system and mission architecture for the exploration of Titan. The proposed robotic platform is capable of flying, rolling and swimming by leveraging the concept of shape-shifting, achieved by attaching and detaching the simple and affordable robotic units, inspired by multi-copters, of which the platform is composed. Thanks to the multi-agent and redundant nature of the system, higher risk exploration activities can be more easily negotiated. In addition, Shapeshifter's morphing capabilities significantly extend the autonomy of the platform, in terms of traveled distance and types of the environment explored, such as caves, cliffs, liquid basins and the cryovolcano Sotra Patera, the main target of our mission on Titan. 
As part of a preliminary feasibility analysis of the concept,  we have developed a simplified, two-agents prototype capable of flying, docking, rolling and un-docking, showcasing the feasibility of key design features and mobility features of the system. From modeling and simulation results we have additionally demonstrated that the morphing capabilities of the robot indeed offer advantages in terms of increased range, as rolling on flat terrains can be up to two times more energy-efficient than flying. 

Future work can proceed in the following directions:
\begin{itemize}
    \item  \textbf{Multi-agent autonomy}: The robustness of a multi-agent system can be better exploited by developing distributed autonomy strategies and strategies that leverage the morphing capabilities of the robot. Examples include a distributed control system for rolling and a motion planner capable of making full use of the morphing abilities of the robot according to the traversability properties of the terrain. 
    \item \textbf{Subsystem definition}: Further refine the identification of power, thermal and communication equipment compatible with the scope of the mission and the requirement of the platform. Detail the requirements for under-liquid operations.
    \item \textbf{Adaptation abilities vs implementation complexity trade-offs}: Define the optimal trade-off between the morphing abilities of the platform and the increased complexity in terms of development and testing. 
\end{itemize}

\paragraph{Note:}
The information presented in this study about the Shapeshifter concept is pre-decisional and is provided for planning and discussion purposes only.

\paragraph{Acknowledgement:}
The research was carried out at the Jet Propulsion Laboratory, California Institute of Technology, under a contract with the National Aeronautics and Space Administration.

\bibliographystyle{plainnat} %
\bibliography{bib/library.bib}

\begin{acronym}
\acro{CoM}{Center of Mass}
\acro{MAV}{Micro Aerial Vehicle}
\acro{TSMP}{Traveling Sales Man Problem}
\acro{VTOL}{Vertical Take Off and Landing}
\acro{ICRA}{International Conference in Robotics and Automation}
\acro{EM}{electro-magnet}
\acro{RTG}{radioisotope termoelectric generator}
\acro{PEM}{permanent electro-magnet}
\acro{IMU}{Inertial Measurement Unit}
\acro{PIXL}{Planetary Instrument for X-Ray Litochemistry}
\acro{LAMP}{Large-scale Autonomous Mapping and Positioning}
\acro{SLAM}{Simultaneous Localization and Mapping}
\acro{MSRG}{Modular Stirling Radioisotope Generator}
\acro{MMRTG}{Multi-Mission Radioisotope Thermoelectric Generator}
\acro{DTN}{disruption tolerant networking}
\acro{TRL}{technology readiness level}
\acro{POMDP}{partially observable Markov decision process}
\acro{MDP}{Markov decision process}
\acro{FIRM}{Feedback-based Information RoadMap}
\acro{SLAP}{Simultaneous Localziation and Planning}
\acro{FMT*}{Fast Marching Tree}
\acro{Dec-POMDP}{Decentralized \ac{POMDP}}
\acro{Caver}{Cave-explorer}
\acro{PM}{programmable magnets}
\end{acronym}

\end{document}